\documentclass[10pt,twocolumn,letterpaper]{article}

\usepackage{iccv}              
\usepackage{algorithm}
\usepackage{algorithmic}
\usepackage{amsmath} 
\usepackage{lineno}
\usepackage{bm}
\usepackage[nice]{nicefrac}
\usepackage{amssymb}
\usepackage{newfloat}
\usepackage{listings}
\usepackage{colortbl}
\usepackage{multirow}
\usepackage{makecell}
\usepackage{dsfont}
\usepackage[accsupp]{axessibility}  
\usepackage{pifont}
\usepackage{cuted} 
\usepackage{wrapfig}
\definecolor{cvprblue}{rgb}{0.21,0.49,0.74}
\definecolor{color3}{RGB}{255, 255, 200}
\definecolor{color2}{RGB}{255, 230, 180}
\definecolor{color1}{RGB}{255, 181, 163}

\newcommand{\best}{\cellcolor{color1}}
\newcommand{\sbest}{\cellcolor{color2}}
\newcommand{\tbest}{\cellcolor{color3}}
\definecolor{iccvblue}{rgb}{0.21,0.49,0.74}
\usepackage[pagebackref,breaklinks,colorlinks,allcolors=iccvblue]{hyperref}

\def\paperID{659} 
\def\confName{ICCV}
\def\confYear{2025}

\title{GS-ID: Illumination Decomposition on Gaussian Splatting via Adaptive Light Aggregation and Diffusion-Guided Material Priors}
\usepackage{hypcap}

\author{
Kang Du$^{1}$ \quad
Zhihao Liang$^{2}$ \quad
Yulin Shen$^{1}$ \quad
Zeyu Wang$^{1,3}\thanks{Corresponding author.}$\\
$^{1}$The Hong Kong University of Science and Technology (Guangzhou)\\
$^{2}$South China University of Technology\\
$^{3}$The Hong Kong University of Science and Technology\\
}
\begin{document}
\maketitle
\begin{strip}
\begin{minipage}{\textwidth}\centering
\vspace{0pt}
\includegraphics[width=0.99\textwidth]{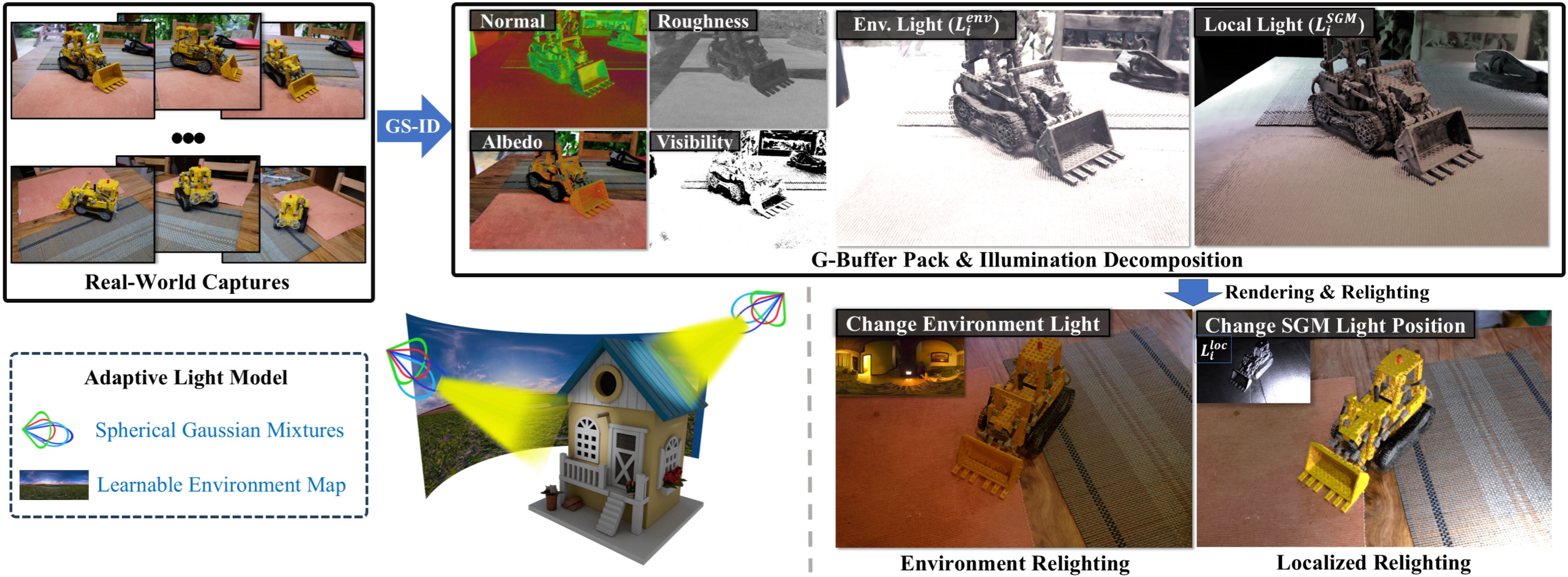}
\captionof{figure}{GS-ID pipeline and relighting results. GS-ID introduces a novel lighting model based on adaptively optimized spherical Gaussian mixtures, enabling precise and editable control of local lighting. 
Combined with per-splat shadow-aware vectors and diffusion-guided material priors, GS-ID achieves state-of-the-art illumination decomposition on 3D Gaussian Splatting.}
\label{fig:teaser}
\end{minipage}
\end{strip}

\begin{abstract}
Gaussian Splatting (GS) has emerged as an effective representation for photorealistic rendering, but the underlying geometry, material, and lighting remain entangled, hindering scene editing.  
Existing GS-based methods struggle to disentangle these components under non-Lambertian conditions, especially in the presence of specularities and shadows.
We propose \textbf{GS-ID}, an end-to-end framework for illumination decomposition that integrates adaptive light aggregation with diffusion-based material priors. 
In addition to a learnable environment map for ambient illumination, we model spatially-varying local lighting using anisotropic spherical Gaussian mixtures (SGMs) that are jointly optimized with scene content. 
To better capture cast shadows, we associate each splat with a learnable unit vector that encodes shadow directions from multiple light sources, further improving material and lighting estimation.
By combining SGMs with intrinsic priors from diffusion models, GS-ID significantly reduces ambiguity in light-material-geometry interactions and achieves state-of-the-art performance on inverse rendering and relighting benchmarks. 
Experiments also demonstrate the effectiveness of GS-ID for downstream applications such as relighting and scene composition.
\end{abstract}
    
\section{Introduction}
\label{sec:intro}
3D Gaussian Splatting (3DGS)~\cite{kerbl20233d} has emerged as a promising 3D representation, offering explicit scene modeling with real-time differentiable rendering capabilities. However, its practical adoption by downstream applications faces a critical limitation: the inherent entanglement of geometry, material, and illumination components during multiview reconstruction, which prevents further editing of individual components. Illumination decomposition on 3DGS has significant value as it can facilitate versatile GS editing using disentangled components, including changing the light and material for scene composition.

There are three critical challenges in achieving effective illumination decomposition on 3DGS.
1) Insufficient light modeling: Existing methods struggle to represent complex lighting environments, often failing to balance global ambient lighting with localized high-frequency effects.
2) Entangled shadows: Shadows arising from intricate light-geometry interactions obscure material estimation and are difficult to isolate.
3) Ill-posed intrinsic estimation: Without priors, decomposing the coupled outputs of 3DGS into geometry, material, and lighting remains highly ambiguous.

Recent methods~\cite{boss2021nerd, chen2021nerv, zhang2023neilf++, jiang2024gaussianshader, wang2021learning, jin2023tensoir, liang2023gsir} model illumination using global lighting representations, such as learnable environment maps or neural light fields. However, they typically assume distant light sources, neglecting near-field effects that are critical for capturing localized specular highlights and accurate material properties.
Emerging solutions like VMINer~\cite{fei2024vminer} and $\text{GS}^3$~\cite{bi2024gs3} incorporate photometric prior knowledge (e.g., predefined light sources, intensity distributions, and angular profiles) into near-field illumination modeling. However, strong parametric assumptions about light configurations restrict their practical applicability, making them dependent on synthetic scenes with calibrated lighting setups.
Shadow modeling is another bottleneck. MII~\cite{zhang2022invrender} employs a learned visibility MLP, but struggles with multiple discrete light sources. Prior-free decomposition under such conditions remains ill-posed, especially in in-the-wild captures.
To resolve geometry-light ambiguity, DN-Splatter~\cite{turkulainen2024dnsplatter} uses pre-trained models to provide pseudo-normal supervision, but it relies on additional sensors (e.g., LiDAR), limiting its applicability.

To address these challenges, we propose GS-ID, a novel end-to-end framework for illumination decomposition on 3DGS. GS-ID integrates an adaptive lighting model, a deshadowing module, and pretrained diffusion-based priors for geometry and material. A customized CUDA-based optimization with deferred rendering further accelerates the decomposition process.
Inspired by production lighting pipelines in tools like Unity~\cite{unity} and Unreal Engine~\cite{unreal}, which combine ambient lighting with localized sources, we design a lighting model that explicitly separates ambient and localized illumination. GS-ID represents ambient light via a learnable environment map and models high-frequency localized lighting using spatially varying spherical Gaussian mixtures (SGMs). These SGMs are initialized on a 3D grid and adaptively aggregated during optimization to capture complex lighting effects (\Cref{subsection:alight_model}).
To handle shadow-induced errors in material estimation, we introduce a deshadowing module that learns per-splat visibility vectors, enabling the network to disentangle shadows caused by multiple unknown light sources (\Cref{subsection:visibility}).
Finally, to resolve ambiguity in joint light-material optimization, we introduce pretrained diffusion priors. For geometry, we incorporate normal priors~\cite{conf/siggraph/0005DGHHLYH24} to stabilize reconstruction. For material, we guide decomposition using albedo and roughness maps from a pretrained diffusion model (\Cref{subsection:priors_from_diffusion}).

\begin{figure*}[t]
\centering
\includegraphics[width=\textwidth]{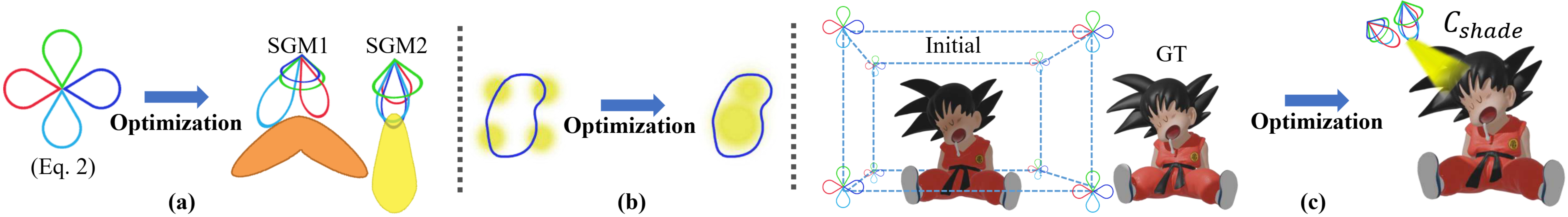}
\caption{Illustration of the SGM optimization process.
(a) A single SGM can adaptively represent different anisotropic illuminations after optimization.
(b) Multiple SGMs can adaptively aggregate to represent complex and irregular highlight regions.
(b) Illustration of the overall adaptive lighting optimization.
The detailed optimization process is available in the supplemental material.
}
\label{fig:lighting_model}
\end{figure*}
This paper makes the following contributions:
\begin{itemize}
\item 
We introduce GS-ID, an end-to-end framework for illumination decomposition, which leverages material priors from pretrained diffusion models to improve joint light and material optimization on 3DGS.
\item 
We propose a novel lighting model supporting adaptive optimization of local and ambient illumination under unknown conditions with a CUDA implementation.
\item We develop a GS-based visibility-aware deshadowing model for efficient shadow approximation caused by multiple light sources, improving material estimation quality.
\end{itemize}

\section{Related Work}
\label{sec:related_work}
\textbf{Geometry Reconstruction.}
Regarding surface reconstruction, some methods~\cite{park2019deepsdf, niemeyer2020differentiable, oechsle2021unisurf, yariv2020multiview, wang2021neus, yariv2021volume} use an MLP to model an implicit field representing the target surface. After training, surfaces are extracted using isosurface extraction algorithms~\cite{lorensen1998marching, ju2002dual}.
Recently, several methods~\cite{huang20242dgs, journals/corr/abs-2404-10772} have achieved fast and high-quality geometry reconstruction of complex scenes based on 3DGS~\cite{kerbl20233d}.
However, these methods often model the resulting appearance given a view direction, neglecting the modeling of physical materials and light transport on the surface.

\textbf{Intrinsic Decomposition.}
To decompose the intrinsics from observations, some monocular methods~\cite{zhu2022irisformer, conf/siggrapha/ZhuLH0ZX0BZT22, kocsis2024intrinsic, conf/siggraph/0005DGHHLYH24} learn from labeled datasets and estimate intrinsics directly from single images.
However, these methods lack multi-view consistency and struggle to tackle out-of-distribution cases.
In contrast, other methods~\cite{bi2020neural, srinivasan2021nerv, zhang2021nerfactor, zhang2022modeling, munkberg2022extracting, jin2023tensoir, jiang2024gaussianshader, liang2023gsir} construct a 3D consistent intrinsic field from multiple observations.
While obtaining impressive results, current methods almost only consider environmental illumination, making precise editing challenging.

\textbf{Lighting Models.} 
Accurate illumination estimation from multi-view observations remains challenging due to inverse rendering ambiguities. Current methodologies primarily follow two paradigms:
1) For unknown illumination conditions, approaches utilize either implicit neural representations (e.g., MLP-based coordinate networks like NeRF variants ~\cite{boss2021nerd, chen2021nerv}) or explicit parametric models (e.g., HDR environment maps in ~\cite{jin2023tensoir, liang2023gsir}). While neural light fields (~\cite{yao2022neilf, zhang2023neilf++}) achieve differentiable surface reconstruction with inter-reflection modeling through MLPs, they trade interpretability for flexibility - excelling at view synthesis but lacking spatially aware editing controls. Explicit directional lighting models enable intuitive environmental map replacement yet fail to resolve localized light interactions.
2) Under known illumination constraints (e.g., calibrated flash images), hybrid decomposition frameworks like VMINER ~\cite{fei2024vminer} leverage controlled captures with/without auxiliary lights to separate near/far-field components within fixed viewpoints. This photometric prior integration addresses partial information scenarios but restricts dynamic scene adaptation.

\begin{figure*}[!t]
\centering
\includegraphics[width=0.95\textwidth]{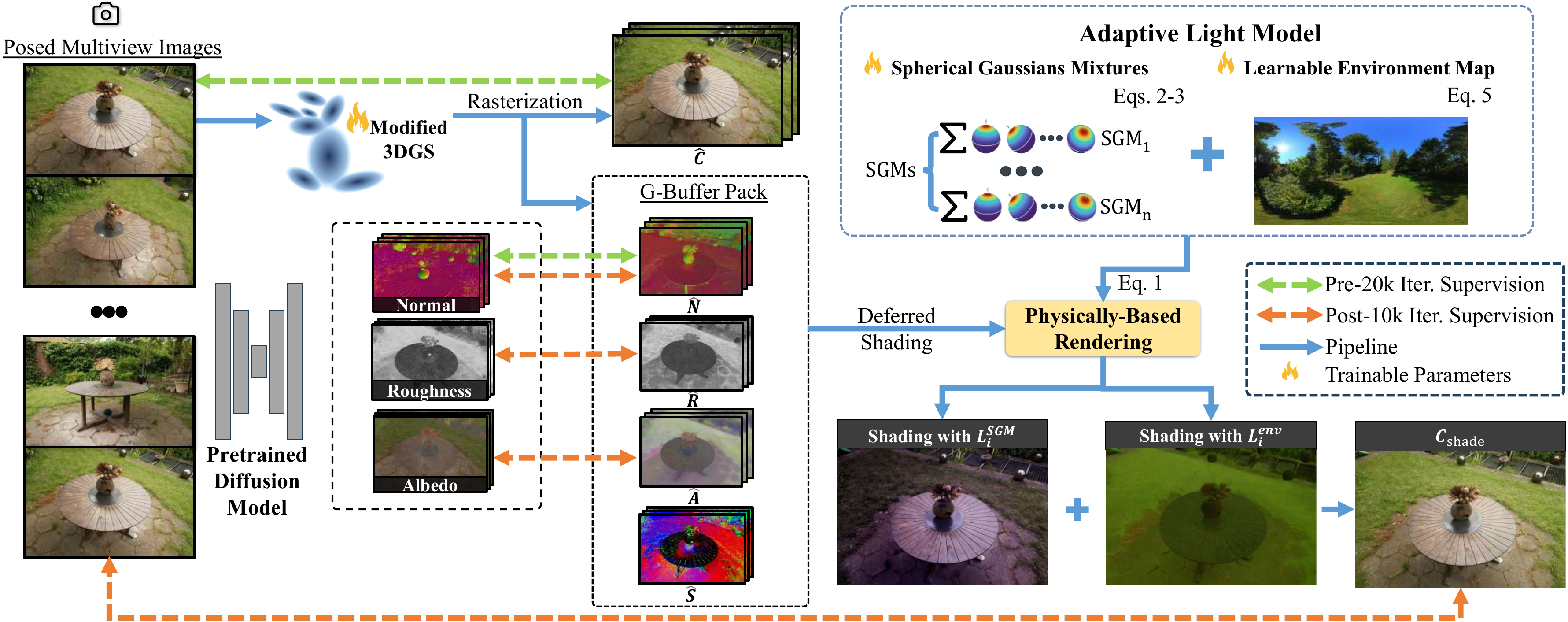} 
\caption{Overview of the GS-ID pipeline. We first reconstruct a coarse 3D Gaussian Splatting (3DGS) scene using normal priors from a diffusion model. Material priors are then incorporated to jointly optimize illumination and intrinsic properties, which are stored as G-Buffer maps for efficient deferred shading. Illumination is represented by an adaptive lighting model composed of spherical Gaussian mixtures (SGMs) and a learnable environment map.}
\label{fig:pipeline}
\end{figure*}

\section{Methodology}
\label{sec:method}
GS-ID uses an adaptive lighting model, a deshadowing model, and diffusion-guided normal and material priors to decompose intrinsic properties from illumination on 3DGS.
We use deferred shading with the G-buffer pack for more efficient rendering.
\Cref{fig:pipeline} shows the complete GS-ID pipeline, and \ref{subsec:implementation} shows the implementation detail.

\subsection{Adaptive Lighting Model}
\label{subsection:alight_model}
Effectively modeling complex lighting conditions in a scene is the first step toward illumination decomposition.
We propose an adaptive lighting model using a set of SGMs to model high-frequency lighting components originating from spatially localized discrete emitters and a learnable environment map to model ambient illumination. The initial SGMs are uniformly placed in a fixed range of $[-3, 3]^3$. Our optimization process aggregates them to model the complex local lighting adaptively.

Using this adaptive lighting model, we formulate incident radiance as: $ L_i = L^\text{SGM}_i + L^\text{env}_i $, where $L^\text{SGM}_i$ accounts for high-frequency effects from discrete emitters and $L^\text{env}_i$ is ambient lighting from distant sources. The rendering equation can be written as:
\begin{equation}
\label{eq:render_eq}
\begin{aligned}
L_o(\bm{x}, \bm{\omega}_o) 
=& \int_{\Omega} L^\text{env}_i(\bm{x}, \bm{\omega}_i) f_r(\bm{\omega}_i, \bm{\omega}_o)
(\bm{\omega}_i \cdot \bm{n}) \text{d}\bm{\omega}_i + \\
& \int_{\Omega} L_i^\text{SGM}(\bm{x}, \bm{\omega}_i) f_r(\bm{\omega}_i, \bm{\omega}_o)
(\bm{\omega}_i \cdot \bm{n}) \text{d}\bm{\omega}_i \\
\approx& L^\text{env}_{o}(\bm{x}, \bm{\omega}_o) + L^\text{SGM}_{o}(\bm{x}, \bm{\omega}_o)*V,
\end{aligned}
\end{equation}
where $\bm{x}$ represents a point in the 3D space and $\bm{n}$ is the surface normal. $f_r$ is the bidirectional reflectance distribution function (BRDF) for physically based rendering~\cite{Nicodemus:65}. $\bm{\omega}_i$ and $\bm{\omega}_o$ refer to the incident and outgoing directions, respectively.
$V$ represents a modulating weight caused by shadow, which is modeled in detail in \Cref{subsection:visibility}.


\subsubsection{SGM-Based Local Lighting}
Local light sources are often accountable for various lighting effects like highlights.
We model complex local lighting using SGMs~\cite{wang2009all}, where each SGM comprises $n_\text{SG}$ individual SGs. The $k$-th SG in an SGM is parameterized by primary emission direction $\bm{b}_k \in \mathbb{S}^2$, sharpness $\lambda_k \in \mathbb{R}^+$, amplitude $\mu_k \in \mathbb{R}^+$, and mixture weights $\bm{w}_k \in \mathbb{R}^3$ controlling RGB chromaticity.
An SGM contains multiple directional lobes represented by the SGs along their primary emission directions, producing a ``spotlight-like'' distribution (\Cref{fig:lighting_model}a).
Multiple SGMs can be adaptively placed in the 3D space and aggregated to model complex light sources that cause irregular highlights.
As shown in \Cref{fig:lighting_model}b, the SGMs can be jointly optimized because of their differentiable parameters:
\begin{equation}
\small
\label{eq:sg_equation}
\begin{aligned}
\text{SGM}(\bm{\omega}_o; \bm{b}, \bm{\lambda}, \bm{\mu}) = \sum_{k=1}^{n_\text{SG}} \mu_k e^{\lambda_k (\bm{\omega}_o \cdot \bm{b}_k - 1)} \cdot \bm{w}_k.
\end{aligned}
\vspace{-0.1cm}
\end{equation}
The output radiance $L^\text{SGM}_{o}$ at the surface point $\bm{x}$ integrates incident illumination through the Cook-Torrance BRDF $f_r$:
\begin{equation}
\small
\label{eq:dir_equation}
\begin{aligned}
L^\text{SGM}_{o}(\bm{x}, \bm{\omega}_o) &= \int_{\Omega} f_r(\bm{\omega}_i, \bm{\omega}_o) L_i^\text{SGM}(\bm{x}, \bm{\omega}_i) (\bm{n} \cdot \bm{\omega}_{i}) \text{d}\bm{\omega}_i\\
&\approx \sum_{j}^{N_\text{light}} \frac{f_{r}^{(j)} \cdot \text{SGM}(\bm{\omega}_o^{(j)}) (\bm{n} \cdot \bm{\omega}_i^{(j)}) * V_j}{|\bm{p}_j - \bm{x}|^2},
\end{aligned}
\vspace{-0.1cm}
\end{equation}
where $N_\text{light}$ SGMs are initialized on a 3D grid and strategically optimized. All SGs in an SGM use the same spatial position $\bm{p}_j$ while maintaining their own SG parameters ($\bm{b}_k,\lambda_k,\mu_k$). The modulating weight caused by shadow, $V_j$, is estimated using a shadow-aware unit vector associated with each splat and each lighting direction, as discussed in \Cref{subsection:visibility} and illustrated in \Cref{fig:lighting_model}. $|\bm{p}_j - \bm{x}|$ is the distance between the surface and the $j$-th light source, modeling illumination decay over distance.
To enhance computational efficiency, we apply progressive pruning of low-energy SGMs with $|\bm{w}_k| < \tau$, as discussed in \Cref{subsec:implementation}.

To ensure that SGM optimization leads to an accurate model of local lighting rather than ambient lighting, we introduce two regularization terms considering the position and value of each SGM light source:
\begin{equation}
\small
\begin{aligned}
\mathcal{L}_\text{pos} & = \sum_{j}^{N_\text{light}} \max(d_{\text{min}}^{(j)}- d_{\text{max}}, 0),\quad
\mathcal{L}_\text{val} = \sum_{j}^{N_\text{light}}\sum_{k}^{n_\text{SG}} \Vert \bm{w}_{jk}\Vert_2\\
&\mathcal{L}_\text{light} =  \lambda_\text{pos} \mathcal{L}_\text{pos} + \lambda_\text{val} \mathcal{L}_\text{val}.
\end{aligned}
\label{eq:loss_lighting}
\vspace{-0.1cm}
\end{equation}
Here \( d_{\text{min}}^{(j)} = \min\limits_{\bm{x}} |\bm{p}_j - \bm{x}| \) represents the minimum distance between the \( j \)-th light position \( \bm{p}_j \) and the surface position \( \bm{x} \) in the scene. The 3D position \( \bm{x} \) in world coordinates is derived by back-projecting the depth value $\hat{D}$ from the depth buffer. The parameter \( d_{\text{max}} \) is a hyperparameter with a default value of 3. The weight \( \bm{w}_{jk} \) is responsible for controlling the contribution of the \( k \)-th SG in the \( j \)-th SGM. We set both \( \lambda_{\text{pos}} \) and \( \lambda_{\text{val}} \) to \( 1 \times 10^{-6} \) by default. These two regularization terms help direct the SGMs toward the surface so that the SGMs can better model complex local lighting.

Our SGM-based lighting model can represent spatially varying illumination through an adaptive mechanism. Compared to existing methods, as illustrated in \Cref{fig:teaser}, our approach can create lighting effects like localized highlight regions and easily support illumination editing.


\subsubsection{Ambient Lighting}
In addition to local lighting, ambient lighting $L^\text{env}_o$ can be reformulated into its diffuse ($L^\text{env}_{o\text{-diff}} $) and specular ($L^\text{env}_{o\text{-spec}}$) components.
We adopt an image-based lighting model and the split-sum approximation~\cite{karis2013real} to handle the intractable integral.
$L^\text{env}_o$ can be represented as:
\begin{equation}
\small
\begin{gathered}
\label{eq:env_diff_spec_approx}
L^\text{env}_{o}(\bm{x}, \bm{\omega}_o) =
L^\text{env}_{o\text{-diff}} + L^\text{env}_{o\text{-spec}},\\
L^\text{env}_{o\text{-diff}} \approx K^\text{env}_\text{diff} I^\text{env}_\text{diff},\quad
K^\text{env}_\text{diff} = (1 - M) \frac{\bm{A}}{\pi},\\
\begin{aligned}
L^\text{env}_{o\text{-spec}}
\approx& \underbrace{
\int_{\Omega} \frac{DFG}{4 (\bm{n} \cdot \bm{\omega}_o)} \, \text{d}\bm{l}
}_{
\text{Environment BRDF }(K^\text{env}_\text{spec})
}
\cdot\underbrace{
\int_{\Omega} 
DL_i(\bm{l}) (\bm{l} \cdot \bm{n}) \, \text{d}\bm{l}.
}_{
\text{Pre-Fil. Env. Map }(I^\text{env}_\text{spec})
}
\end{aligned}
\end{gathered}
\vspace{-0.1cm}
\end{equation}
$K^\text{env}_\text{spec}$ can be quickly accessed in precomputed lookup tables. $I^\text{env}_\text{diff}$ and $I^\text{env}_\text{spec}$ are embedded within a learnable environment map. 

\begin{figure*}[t]
\centering
\includegraphics[width=\linewidth]{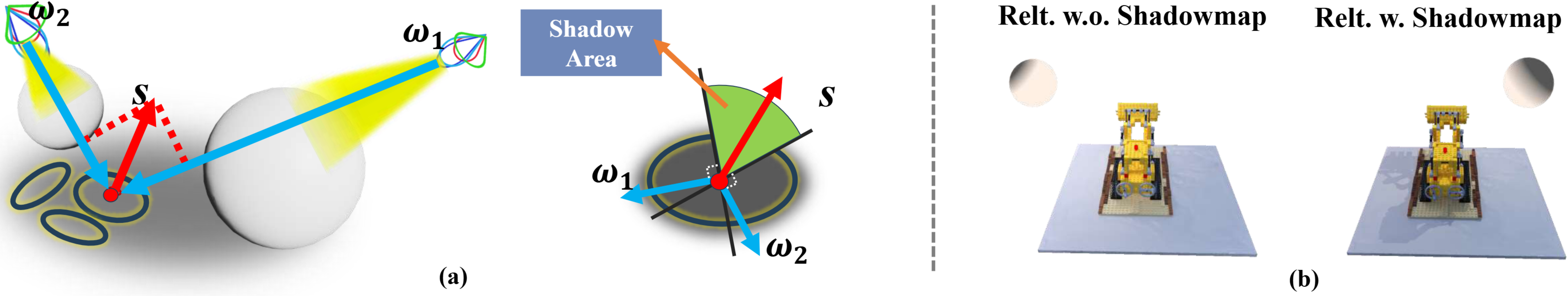}
\vspace{-0.6cm}
\caption{Illustration of our proposed deshadowing model. The lighting directions are denoted by $\bm{\omega}_1$ and $\bm{\omega}_2$. For each Gaussian splat, we optimize a unit vector $\bm{s}$ to represent the dominant direction in which a light casts a shadow on it.
}
\label{fig:fig_deshadowing}
\end{figure*}

\subsection{Deshadowing Model}\label{subsection:visibility}
During reconstruction, cast shadows are often baked into the albedo, leading to inaccurate material estimation. Existing methods like ray tracing or offline baking~\cite{liang2023gsir} are non-differentiable and computationally expensive, limiting scalability.
To address this during training, we assign each 3DGS primitive a learnable unit vector ${\bm{s}} \in \mathbb{S}^2$ that captures the dominant shadow direction under multiple lights. These vectors are alpha-blended~\cite{kerbl20233d} into a screen-space shadow field and integrated into the G-buffer, enabling efficient, differentiable shadow prediction without explicit ray tracing (see \Cref{fig:fig_deshadowing}a).
This deshadowing mechanism is used only during training. In inference, standard shadow mapping techniques are applied for relighting (see \Cref{fig:fig_deshadowing}b).

The visibility of incident light from direction $\bm{\omega}_i^{(j)}$ is estimated as:
\begin{equation}
V_j = \sigma\left(\alpha \cdot \hat{\bm{S}} \cdot \bm{\omega}_i^{(j)} + \beta\right),
\vspace{-0.1cm}
\label{eq:visibility}
\end{equation}
where $\hat{\bm{S}}$ is a screen-space shadow vector field derived from alpha-blending the per-primitive vectors $\bm{s}$ in the G-Buffer. 
Here, $\sigma(\cdot)$ denotes the sigmoid function. 
The hyperparameters $\alpha$ and $\beta$ modulate the sharpness and baseline level of shadow effects, respectively, with default values $\alpha = 8$ and $\beta = 10^{-3}$.
This differentiable formulation facilitates shadow-material disentanglement during training, as visualized in \Cref{fig:real_world} and validated through ablation in \Cref{fig:ablation_near}. 

\begin{figure*}[t]
\centering
\includegraphics[width=0.99\textwidth]{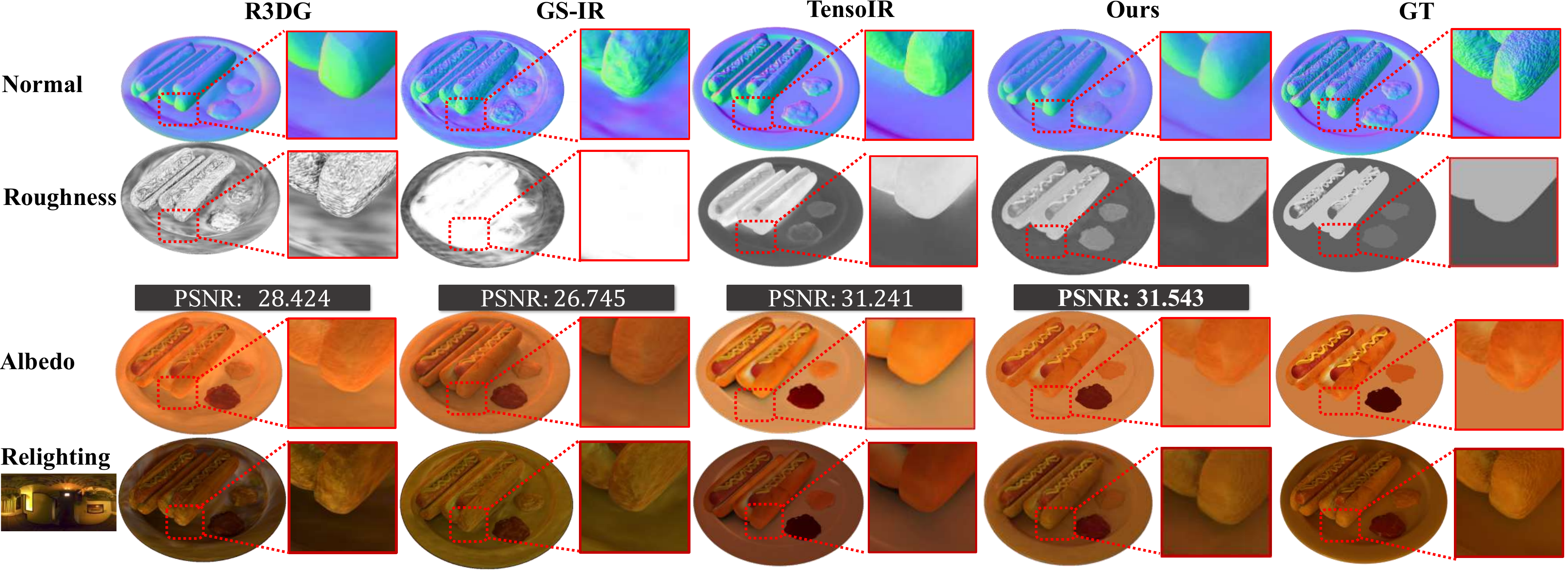}
\caption{Qualitative evaluation on the TensoIR dataset.
GS-ID disentangles shadows effectively and yields more accurate albedo, normal, and roughness maps. For albedo and roughness results, we follow \cite{zhang2021nerfactor} and scale each RGB channel by a global scalar.}
\label{fig:material_decom}
\end{figure*}
\begin{table*}[!t]
\centering
\resizebox{\textwidth}{!}{
    \begin{tabular}{@{}clcc ccc ccc ccc c}
    \toprule
    & \multirow[b]{2}{*}{Method} &
    \multirow{2}{*}[-6pt]{\makecell[t]{Roughness\\MSE $\downarrow$}} &
    \multirow{2}{*}[-6pt]{\makecell[t]{Normal\\MAE $\downarrow$}} &
    \multicolumn{3}{c}{Albedo} &
    \multicolumn{3}{c}{Novel View Synthesis} &
    \multicolumn{3}{c}{Relighting} &
    \multirow{2}{*}[-6pt]{\makecell[t]{Training \\ Time}}\\
    \cmidrule(lr){5-7}\cmidrule(lr){8-10}\cmidrule(lr){11-13}
    & & & & 
    PSNR $\uparrow$ & SSIM $\uparrow$ & LPIPS $\downarrow$ & 
    PSNR $\uparrow$ & SSIM $\uparrow$ & LPIPS $\downarrow$ & 
    PSNR $\uparrow$ & SSIM $\uparrow$ & LPIPS $\downarrow$  \\
    \midrule
    \multirow[]{4}{*}{\rotatebox{90}{NeRF}}
    & InvRender~\cite{zhang2022modeling} &
     \sbest0.008 & 5.074 & 
    27.34 & 0.933 & 0.100 &
    27.37 & 0.934 & 0.089 &
    23.97 & 0.901 & 0.101 &
    15 hrs \\
    & NVDiffrec$^\dag$~\cite{munkberg2022extracting} &
     0.010 & 6.078 & 
    29.17 & 0.908 & 0.115 &
    30.70 & 0.962 & 0.052 &
    19.88 & 0.879 & 0.102 &
    1.2 hrs \\
    & TensoIR~\cite{jin2023tensoir}&
     0.013 & \best 4.100 & 
    29.28 & \sbest0.950 & \tbest0.085 &
    35.09 & \tbest0.976 & 0.040 &
    \sbest28.58 & \sbest0.944 & \tbest0.081 &
    3.9 hrs\\
    \midrule
    \multirow[]{4}{*}{\rotatebox{90}{3DGS}}
    & GSshader$^\dag$~\cite{jiang2024gaussianshader}&
     0.065 &  6.647 & 
    18.59 & 0.876 & 0.092 &
    19.99 & 0.891 & 0.089 &
    22.42 & 0.872 & 0.103 &
    \sbest35 min \\
    & RelightGS$^\dag$~\cite{R3DG2023}&
     0.016 & 6.078 & 
     \tbest29.47 & 0.930 & 0.107 &
    \sbest 37.57 & \sbest0.983 & \best 0.020 &
    \sbest 24.41 & \tbest0.890 & \sbest 0.100 &
    {41 min} \\
    & GS-IR$^\dag$~\cite{liang2023gsir}&
     0.027 &  \tbest4.947 & 
    \sbest30.29 & \tbest0.941 & \sbest0.084 &
    \tbest35.33 & 0.974 & \sbest0.027 &
    24.37 & 0.885 & 0.096 &
    \best26 min \\
    & Ours$^\dag$ &
    \best 0.007 & \sbest 4.602 & 
    \best 33.49 & \best 0.952 & \best 0.079 &
    \best 39.13 & \best 0.984 & \best 0.020 &
    \best 28.69 & \best 0.947 & \best 0.075 &
    \tbest 40 min \\
    \bottomrule
    \end{tabular}
}
\caption{
Quantitative evaluation on the TensoIR Synthetic dataset.
Real-time methods are marked with $^\dag$. Notably, we adopt the same evaluation metrics as used in TensoIR. To compute the average variation, we utilize the ground truth albedo and roughness.
}
\label{tab:synthetic_comparison}
\end{table*}


\begin{table}[t]
\centering
\resizebox{0.99\columnwidth}{!}{
\begin{tabular}{@{}l|ccc|ccc}
\toprule
\multirow{2}{*}{Method} & \multicolumn{3}{c@{}|}{Albedo} & \multicolumn{3}{c@{}}{Novel View Synthesis} \\ 
& PSNR$\uparrow$ & SSIM$\uparrow$ & LIPPS$\downarrow$ &
PSNR$\uparrow$ & SSIM$\uparrow$ & LIPPS$\downarrow$ 
\\
\hline
GSshader$^\dag$~\cite{jiang2024gaussianshader}&
25.375&	0.912&	0.071 & 36.248&0.967&0.043 \\
GS-IR$^\dag$~\cite{liang2023gsir}&
25.471&	0.926&	0.068 & 36.858&0.971&0.041
\\
RelightGS$^\dag$~\cite{R3DG2023}& 27.223& 0.953&	0.057 & 40.228&	\textbf{0.990}&	\textbf{0.012}
\\
Ours ($3^3$ SGMs)$^\dag$ &
\textbf{29.133} & \textbf{0.954} & \textbf{0.056}  &
\textbf{40.412} & 0.988 & \textbf{0.012} 
\\
\bottomrule
\end{tabular}
}
\caption{Quantitative comparison on the ADT dataset.}
\label{table:adt}
\end{table}

\subsection{Diffusion-Guided Priors}
\subsubsection{3DGS Reconstruction with Normal Priors}
\label{subsection:priors_from_diffusion}
We observe that 3DGS reconstruction sometimes mistakenly interprets glossy regions as holes, as shown in~\Cref{fig:ablation_normal}. To address this issue, we incorporate priors from a monocular geometric estimator to improve the output geometric structures.
Specifically, we leverage a pretrained diffusion model RGB$\leftrightarrow$X~\cite{conf/siggraph/0005DGHHLYH24} to provide normal supervision.
The supervision loss $\mathcal{L}_{\text{base}}$ is defined as:
\begin{equation}
\label{eq:2dgs_loss}
\begin{aligned}
\mathcal{L}_{\text{base}} &= \mathcal{L}_{\text{color}} + \lambda_{\bm{n}} \mathcal{L}_{\bm{n}},\\
\mathcal{L}_{\bm{n}} &= \sum_{\hat{\bm{n}} \subset \hat{\bm{N}}} \mathds{1} \left(
1 - \hat{\bm{n}}^T \bm{n}
\right),
\end{aligned}
\end{equation}
where $\mathcal{L}_\text{color}$ is an RGB reconstruction loss that combines the L1 loss with D-SSIM from 3DGS.
We use RGB$\leftrightarrow$X to estimate the normal vector $\bm{n}$, treating it as a pseudo ground truth (pseudo-GT) normal, to supervise our rendered normal $\hat{\bm{n}}$ in the G-buffer pack.
We propose a depth-masked weighting scheme to avoid unreliable priors from distant areas.
We assign a weight of 0 to pixels with depth values exceeding a default threshold of 0.8 and a weight of 1 otherwise, which we denote as $\mathds{1}$ in~\Cref{eq:2dgs_loss}. The hyperparameter \( \lambda_{\bm{n}} \) is set to a default value of 0.05.
By incorporating normal priors, we improve the accuracy of normal estimation and address geometry reconstruction challenges in textureless regions, providing robust normal estimation.

\subsubsection{Light Optimization with Material Priors}

To reduce the inherent ambiguity in joint light optimization, we leverage pseudo-GT material attributes derived from RGB$\leftrightarrow X$~\cite{conf/siggraph/0005DGHHLYH24}:
\begin{equation}
\label{eq:loss_material}
\begin{aligned}
\mathcal{L}_{\text{material}}
&= \lambda_{R}  L_2{(R, \widehat{R})}
+ \lambda_{\bm{A}} L_2{(\bm{A}, \widehat{\bm{A}})},
\end{aligned}
\vspace{-0.1cm}
\end{equation}
where $\widehat{R}$ and $\widehat{\bm{A}}$ denote estimated roughness and RGB albedo maps from the G-buffer pack, while $R$ and $\bm{A}$ represent their corresponding pseudo-GT values. We omit the metallic term as it shows a weak impact on lighting effects. The hyperparameters $\lambda_R$ and $\lambda_{\bm{A}}$ are set to default values of 0.1 and 1.0, respectively.
Combining \Cref{eq:2dgs_loss,eq:loss_lighting,eq:loss_material}, the total loss is:
\begin{equation}
\label{eq:s2_loss}
\begin{aligned}
\mathcal{L}_{\text{total}} = & \mathcal{L}_{\text{base}} + \mathcal{L}_\text{light} +\mathcal{L}_{\text{material}},
\end{aligned}
\vspace{-0.1cm}
\end{equation} where $\mathcal{L}_{\text{light}}$ and $\mathcal{L}_{\text{material}}$ take effect after 10k iterations after an initial scene is reconstructed.

\begin{figure*}[t]
\centering
\includegraphics[width=\textwidth]{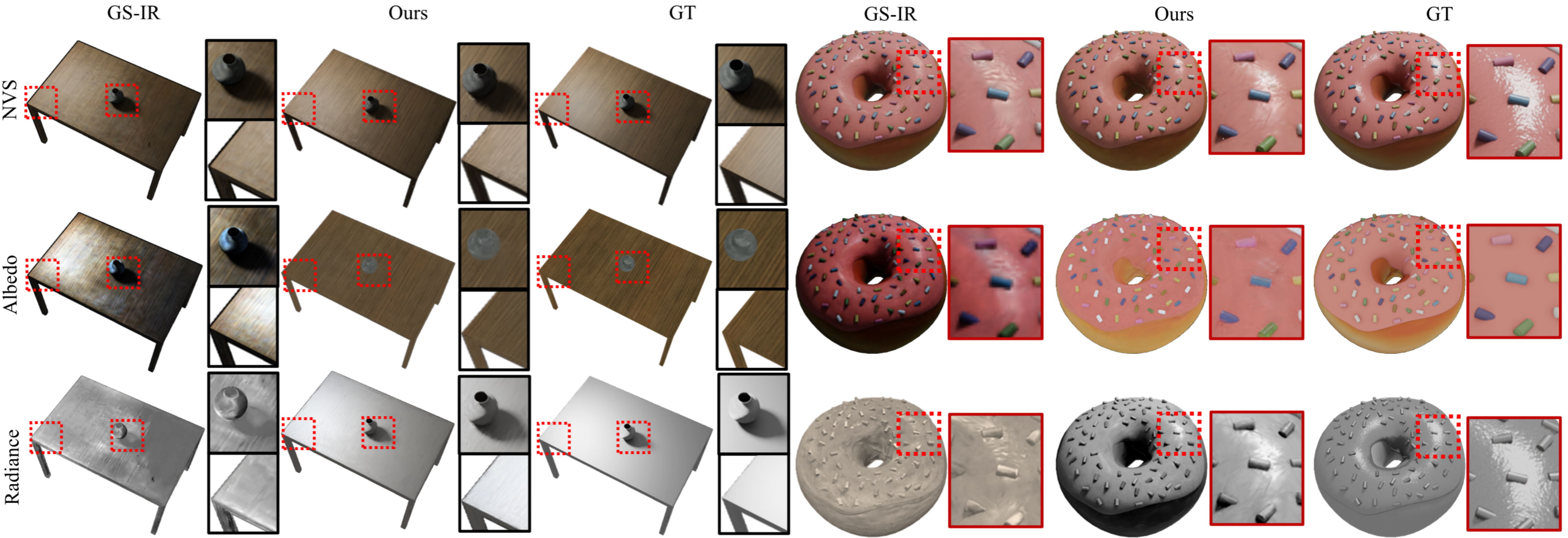}
\caption{Qualitative comparison on the ADT dataset.
GS-ID separates different illumination components and removes specular highlights and shadow artifacts from albedo estimation, significantly outperforming existing solutions under unknown lighting conditions.}
\label{fig:compare_near}
\end{figure*}

\begin{figure*}[t]
\centering
\includegraphics[width=1.0\textwidth]{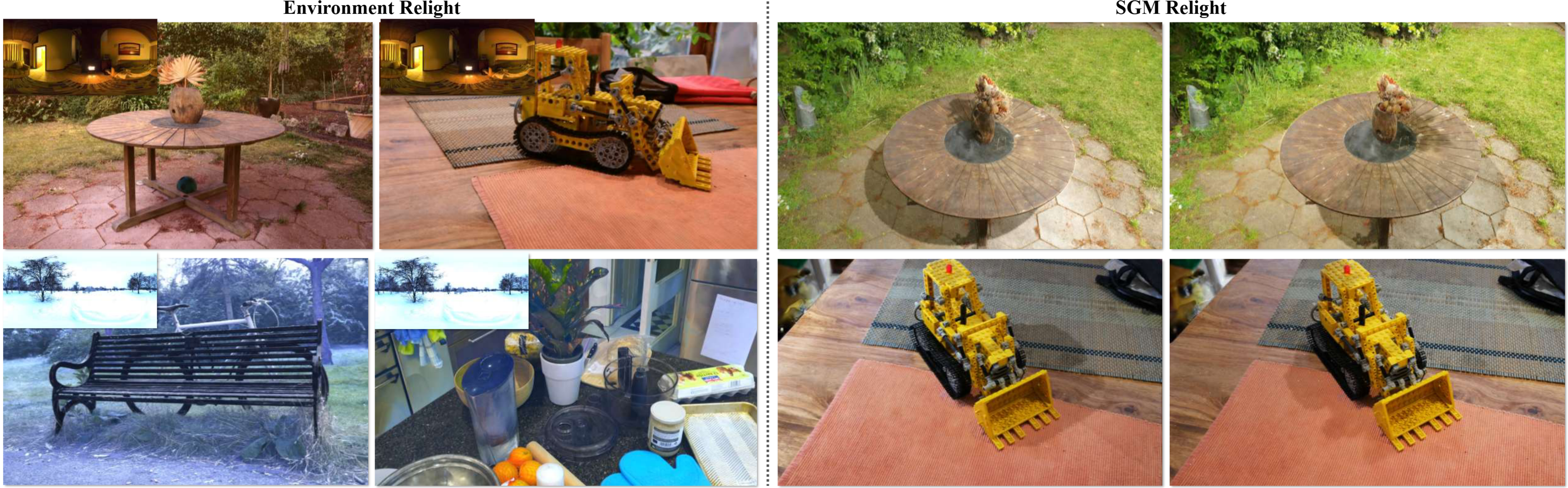}
\caption{Relighting results on the Mip-NeRF 360 dataset. Our adaptive lighting model supports both environment relighting and SGM relighting with shadow maps. Remarkably, relighting is achieved solely by modifying light parameters, with no need for retraining.}

\label{fig:real_world}
\end{figure*}

\section{Experiments} \label{sec:exp}
\subsection{Implementation Details} \label{subsec:implementation}
We implement GS-ID and conduct experiments on an NVIDIA RTX 4090 GPU. The pipeline begins by adapting the original 3DGS method~\cite{kerbl20233d} for 20k iterations to reconstruct robust scene geometry, assisted by normal priors.
For illumination decomposition, we initialize $N_\text{light} = M^3$ localized spherical Gaussian mixtures (SGMs) uniformly distributed within the scene’s axis-aligned bounding box (AABB), where $M$ controls the spatial resolution. We empirically set $M=3$ to balance expressiveness and computational efficiency. Each SGM contains $n_\text{SG}=16$ spherical Gaussians.
The position $\bm{p}_j$ of each SGM is defined as:
\begin{equation}
\small
\bm{p}_j = \bm{c}_{\text{min}} + \left(\frac{j_x}{M{-}1}, \frac{j_y}{M{-}1}, \frac{j_z}{M{-}1}\right) \odot (\bm{c}_{\text{max}} - \bm{c}_{\text{min}}),
\vspace{-0.1cm}
\end{equation}
\noindent where $\bm{c}_{\text{min}}$ and $\bm{c}_{\text{max}}$ are the AABB bounds (default: $[-3, 3]$), and $j_x, j_y, j_z$ index the grid.

Other hyperparameters, including the shadow vector blending parameters $(\alpha, \beta) = (8, 10^{-3})$, are selected empirically to ensure stable gradient behavior. Crucially, our framework includes light regularization terms—both value- and position-based—that consistently guide the optimization to convergence across different initializations and hyperparameter choices, reducing the sensitivity to manual tuning.
To improve training efficiency, we employ an energy-aware pruning strategy that adaptively disables low-activation light sources during optimization:
\begin{equation}
\bm{w}_{\text{max}} = \underset{i,j,k}{\max}~\bm{w}_{ijk}^{(t)}, \quad
\tau^{(t)} = \bm{w}_{\text{max}} + \ln(\delta),
\end{equation}
\noindent where $i \in [1, N_\text{light}]$, $j \in [1, n_\text{SG}]$, $k$ denotes the RGB channels, and $\delta{=}10^{-3}$ by default. SGMs with all weights below the threshold $\tau$ are progressively pruned throughout training.

The memory complexity of screen-space lighting evaluation, $O(HW N_\text{light} n_\text{SG})$, is addressed via customized CUDA kernels using chunked processing and G-Buffer–aware computation to reduce peak GPU memory.
After light initialization and pruning, joint optimization of geometry, lighting, and material proceeds for an additional 10k iterations under the supervision of pretrained diffusion priors. The full pipeline achieves real-time rendering at 60 fps with approximately $1.2\times$ GPU memory usage and $1.5\times$ longer training time compared to vanilla 3DGS.

\subsection{Datasets and Metrics}
We evaluate GS-ID on three complementary datasets to comprehensively assess illumination decomposition performance across both synthetic and real-world scenarios.
Our primary benchmark is the TensoIR Synthetic dataset~\cite{jin2023tensoir}, which provides four object-centric scenes with ground-truth albedo and roughness maps. We further enhance this dataset by rendering additional screen-space roughness maps to supplement its existing intrinsic ground truth.
To validate real-world performance, we adopt two sources:  
\textbf{(1)} A curated relighting dataset of four high-fidelity scanned objects from the Aria Digital Twin (ADT) repository~\cite{pan2023aria}, which provides high-resolution geometry and material under varied lighting;  
\textbf{(2)} Nine unbounded real-world scenes from Mip-NeRF 360~\cite{barron2022mip}, covering diverse indoor and outdoor environments with complex, spatially-varying illumination.
For evaluation, we use a combination of standard novel view synthesis (NVS) metrics—PSNR, SSIM, and LPIPS—as well as mean angular error (MAE) of surface normals to measure the fidelity of geometric reconstruction.
To ensure a fair comparison, we slightly adjust the rendering equation in \Cref{eq:render_eq} to:
\begin{equation}
L^{\text{radiance}}_{o} = \frac{L^{\text{env}}_{o}(\bm{x}, \bm{\omega}_o)}{\hat{\bm{A}}} + L^{\text{SGM}}_{o}(\bm{x}, \bm{\omega}_o) \cdot V,
\end{equation}
where $\hat{\bm{A}}$ denotes the albedo from the G-Buffer and $V$ represents the visibility term. This adjustment aligns our prediction with radiance-only baselines for consistent evaluation.
Additional results on more scenes and evaluation details are provided in the supplementary material.

\subsection{Comparative Analysis}
\label{subsec:comparison}
\textbf{Illumination Decomposition Analysis on TensoIR.}
Quantitative results on the TensoIR dataset (Table~\ref{tab:synthetic_comparison}) demonstrate the comprehensive advantages of our method. We achieve a PSNR of 39.13 dB for novel view synthesis (NVS), surpassing the second-best method R3DG by 1.56 dB, and reduce the surface normal estimation error by 24.2\% (4.602 vs. 6.078 MAE). While TensoIR achieves slightly lower normal error due to its spatially continuous MLP branch, our GS-based design enables more efficient joint optimization, leading to improved albedo, roughness, and overall relighting performance. Furthermore, our method reduces training time by 87\% and significantly lowers memory consumption. It also outperforms all GS-based baselines in both normal estimation and relighting quality.

\textbf{Illumination Decomposition Analysis on ADT.}
Our experiments on the ADT dataset focus explicitly on evaluating material (albedo) and illumination (radiant intensity) decomposition accuracy. As shown in \Cref{fig:compare_near}, our framework robustly separates diffuse reflectance from high-frequency lighting effects, achieving cleaner albedo maps and physically consistent radiance fields compared to baselines—a direct outcome of our adaptive light model. Existing methods, by contrast, exhibit residual highlights contaminating their radiance predictions or oversmoothed/altered textures degrading their estimated albedo (e.g., specular surfaces). Quantitatively (\Cref{table:adt}), we outperform alternatives across both tasks, with particularly significant margins (+12\% PSNR) for NVS due to faithful disentanglement of these components.

\textbf{Illumination Decomposition Analysis on Real-World Dataset.}
On the unbounded real-world Mip-NeRF 360 dataset, our framework demonstrates robust performance under challenging natural illumination. As shown in \Cref{fig:real_world}, our method supports both far-field and near-field relighting with high visual fidelity.

\subsection{Ablation Study}
We conduct an ablation study to evaluate the effectiveness of the introduced diffusion priors, the proposed lighting model, and our CUDA deferred shading scheme.

\begin{figure}[t]
\centering
\includegraphics[width=0.95\columnwidth]{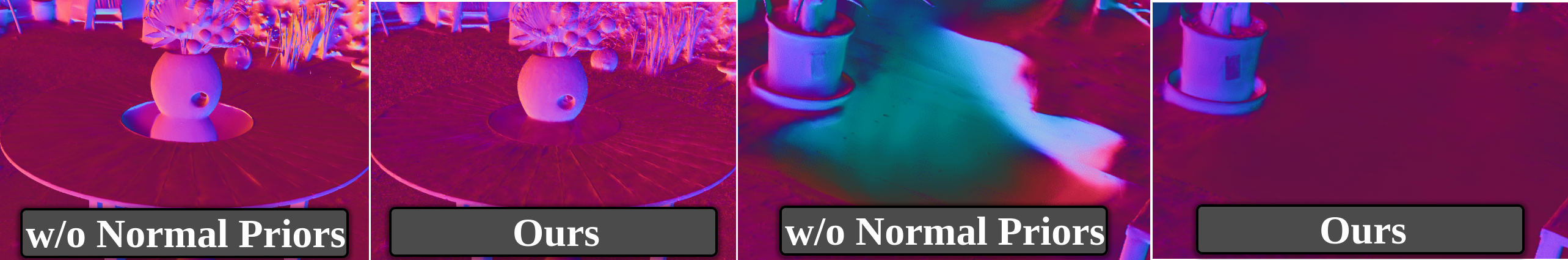}
\caption{Effects of normal priors. Normal priors prevent areas with specular reflections from being misestimated as holes, significantly improving normal estimation accuracy.}
\label{fig:ablation_normal}
\end{figure}

\textbf{Analysis of Diffusion Priors.}
To validate the efficacy of diffusion-derived priors, we categorize them into two distinct components: normal priors and material priors. The quantitative comparisons in \Cref{table:ablation} show that both priors improve the results of NVS and illumination decomposition.
\Cref{fig:ablation_normal} shows that normal prior elimination leads to holistic misestimation of specular areas.
Our extensive validation confirms that the proposed hyperparameter configuration achieves optimal performance.

\begin{figure}[t]
\centering
\includegraphics[width=\columnwidth]{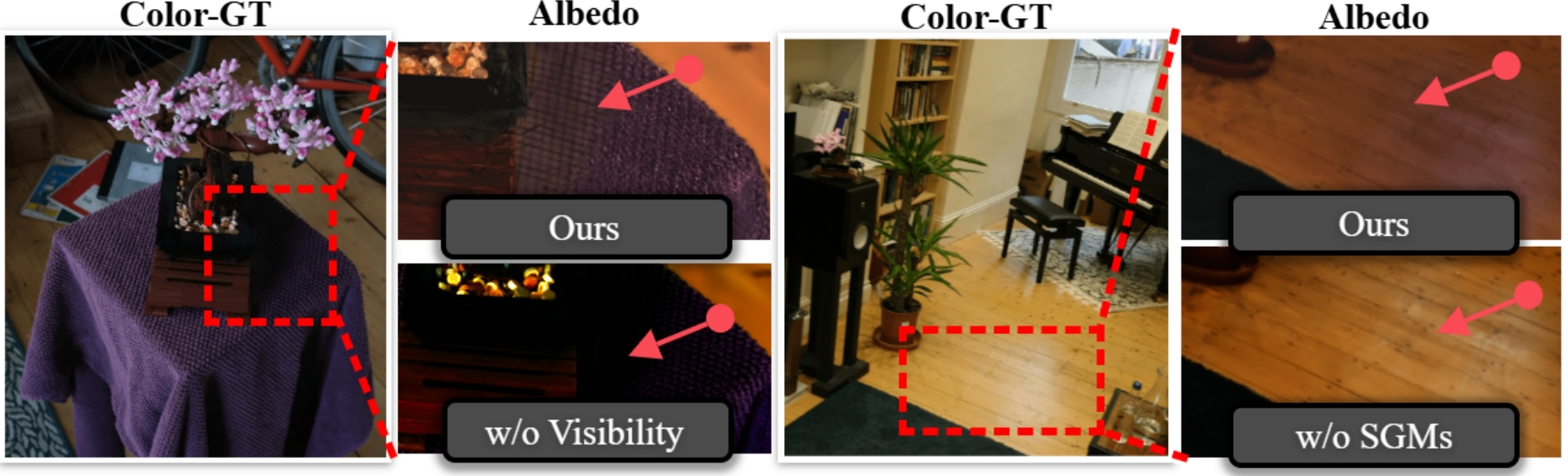}
\caption{Effects of lighting components. Left: Shadow field estimation removes shadow artifacts from albedo estimation. Right: Adaptive SGMs enable accurate specular highlight reconstruction in albedo estimation.}
\label{fig:ablation_near}
\end{figure}

\textbf{Analysis of Lighting Model.}
GS-ID employs an adaptive lighting model that combines parametric SGMs with an environment map to capture complex illumination. We conduct ablations by disabling SGMs and removing the EnvMap. As shown in \Cref{table:ablation}, combining both components improves albedo and NVS accuracy, with SGMs offering a larger PSNR gain for NVS (38.18 dB vs. 36.67 dB). \Cref{fig:ablation_near} shows SGMs better capture high-frequency specularities, while the absence of a shadow field hinders shadow removal, reducing albedo accuracy by 3.213 dB.

\textbf{Analysis of Shading Schemes.}
Forward shading with PBR requires substantial computational resources, as it computes the rendering equation for each primitive. This issue is exacerbated when rendering with Gaussian Splatting, where millions of GS points are typically present in a scene. We compared the training times between forward and deferred shading across various scenes from the Mip-NeRF 360 dataset~\cite{barron2022mip}. The results (\Cref{table:def_acc}) demonstrate that deferred shading accelerates training, with faster scaling with the number of points, achieving up to 4x speedup. Furthermore, our CUDA implementation of differentiable rendering enables saving over 40\% storage space.

\textbf{Discussion and Limitations.}\label{subsec:limitation} 
Our work has achieved good performance in illumination decomposition, but some directions are worth further exploration. We found that natural materials and geometry are often isotropic, while our current 3DGS representation is anisotropic, creating extra Gaussian shapes. We hope to develop an isotropic representation to save storage space in the future.



\begin{table}[t]
\centering
\resizebox{1.0\columnwidth}{!}{
\begin{tabular}{l|c|c c c|c c c}
\hline
\multirow{2}{*}{Method} & Normal &
\multicolumn{3}{c|}{Albedo} &
\multicolumn{3}{c}{Novel View Synthesis}  \\
&MAE$\downarrow$ &
PSNR$\uparrow$ & SSIM$\uparrow$ & LPIPS$\downarrow$ &
PSNR$\uparrow$ & SSIM$\uparrow$ & LPIPS$\downarrow$ \\
\hline
w/o normal & 5.187 &
32.140&0.941&0.089 &
37.962 & 0.978 & 0.028 \\
w/o material & 4.867 &
29.585 & 0.935&0.096 &
38.189 & 0.975  & 0.027 \\
w/o all priors & 5.286 &
29.101&0.921&0.099& 
37.950 & 0.962 & 0.030 \\
w/o SGMs & 5.916 &
26.725&0.911&0.096& 
36.670 & 0.953 & 0.031 \\
w/o EnvMap & 4.701 &
32.320&0.942&0.090&
38.180 & 0.975 & 0.028 \\
w/o shadow field & 4.751 &
30.280&0.941&0.091&
37.580 & 0.974 & 0.029 \\
single SGM & 4.616 &
32.191&0.950&0.089& 
38.670 & 0.982 & 0.025 \\
forward & 4.660 &
32.688 & 0.951 & 0.088 &
39.015 & 0.980 & 0.024 \\
\hline
Ours & \textbf{4.602} &
\textbf{33.493} & \textbf{0.952} & \textbf{0.079} & 
\textbf{39.130}&\textbf{0.984}&\textbf{0.020}\\
\hline
\end{tabular}}
\caption{
Ablation study on the TensoIR Synthetic dataset. The results show the effects of priors from diffusion models, our proposed deshadowing model, and the adaptive lighting model.
}
\label{table:ablation}
\end{table}

\begin{table}[t]
\centering
\resizebox{0.99\columnwidth}{!}{
\begin{tabular}{l|c|c|c|c|c}
\hline
\multirow{2}{*}{Scene} & \multirow{2}{*}{Points} & \multicolumn{2}{c|}{Training Time$\downarrow$ (w. CUDA)} & \multicolumn{2}{c}{GPU Memory$\downarrow$ (Def. Shading)} \\
\cline{3-6}
 &  & Fwd. Shading & Def. Shading & w/o CUDA & w. CUDA \\
\hline
room & 1.23M & 82min & 62min & 28.2G & 18.5G \\
kitchen & 1.31M & 88min & 66min & 29.3G & 18.8G \\
garden & 3.55M & 299min & 68min & 36.8G & 21.4G \\
bicycle & 3.84M & 305min & 70min & 38.9G & 21.9G \\
\hline
\end{tabular}}
\caption{
Comparison of shading schemes on the Mip-NeRF 360~\cite{barron2022mip} dataset.
Deferred rendering can achieve up to a 4× acceleration in complex scenes, with consistent speedup. Our CUDA implementation can save over 40\% GPU memory.
}
\label{table:def_acc}
\end{table}

\section{Conclusion}
We propose GS-ID, an end-to-end framework for illumination decomposition on 3D Gaussian Splatting. It integrates an adaptive lighting model, a deshadowing module, and geometry/material priors extracted from pretrained diffusion models. GS-ID further incorporates a customized CUDA-based optimization pipeline and deferred shading to improve convergence and efficiency. Experiments show that GS-ID surpasses existing methods in both illumination and intrinsic decomposition. Moreover, it enables high-quality, user-controllable light editing and scene composition, supporting diverse downstream applications.

\clearpage
\noindent\textbf{Acknowledgments.}
This research was supported by Meituan Academy of Robotics Shenzhen and HKUST(GZ) LASERi Seed Grant.

{
    \small
    \bibliographystyle{ieeenat_fullname}
    \bibliography{main}
}

\clearpage
\setcounter{page}{1}
\maketitlesupplementary
This supplementary material provides a detailed description of our method's implementation, followed by additional results from various datasets. We then introduce a new experiment to compare the illumination decomposition results under simple and complex light sources, reinforcing the arguments presented in the main paper. Finally, we present further application results. The supplementary material video files include additional videos demonstrating our work.

\section{Implementation Details}
We implement GS-ID using CUDA extensions and modify 3DGS to output G-buffer properties, including albedo, roughness, metallic, normal, and depth, visibility vector.
\subsection{Representation}
In the vanilla 3DGS, each 2D Gaussian utilizes learnable parameters $\mathcal{T} = \{\bm{p}, \bm{s}, \bm{q} \}$ and $\mathcal{C} = \{\alpha, \bm{f}_{c}, \bm{N}, \bm{D} \}$ to describe its geometric properties and volumetric appearance, respectively. 
Here, $\bm{p}$ denotes the position vector, $\bm{s}$ denotes the scaling vector, $\bm{q}$ denotes the unit quaternion for rotation, $\alpha$ denotes the opacity,  $\bm{N}$ denotes the normal,  $\bm{D}$ denotes the depth, and $\bm{f}_{c}$ denotes the spherical harmonics (SH) coefficients for view-dependent color.
In GS-ID, we extend $\mathcal{C}$ to $\{\alpha, \bm{f}_{c},\bm{N}, \bm{D}, \bm{A}, R, M\}$ to describe the material properties of the 2D Gaussian, and $\bm{A}, R, M$ denote Albedo, Roughness, and Metallic values. Additionally, on each pixel in screen space, we can calculate the world space positions $\bm{P}$ by rasterizing $\bm{D}$ with the camera transformation matrix.
\subsection{Training Details}
Our training pipeline consists of two stages: pre-20k and post-10k iterations.

\noindent\textbf{Pre-20k Stage.}
The primary goal of this stage is to establish a stable geometric structure, particularly accurate surface normals and point positions for rasterization. We run 20k iterations using a modified 3D Gaussian Splatting (3DGS) framework, where surface normals predicted by a pretrained diffusion model serve as supervision. This stage ensures geometric consistency before introducing illumination modeling.

\noindent\textbf{Post-10k Stage.}
In this stage, we freeze the Gaussian point positions and use the modified 3DGS to predict G-Buffer components, including albedo, roughness, metallic, and a unit shadow direction vector. These outputs are supervised by corresponding maps generated from the diffusion model. Subsequently, we optimize our illumination model, which comprises a set of learnable spherical Gaussian mixtures (SGMs) and a trainable environment light. We introduce a lighting regularization term and train for an additional 10k iterations. The final supervision is provided by comparing rendered RGB images against ground truth.

\begin{equation}
\label{eq:brdf:supp}
\begin{aligned}
f_r(\bm{\omega}_i, \bm{\omega}_o) &= 
\underbrace{
(1 - M) \frac{\bm{A}}{\pi}
}_{\text{diffuse component}} +
\underbrace{
\frac{DFG}{4(\bm{n} \cdot \bm{\omega}_i)(\bm{n} \cdot \bm{\omega}_o)}
}_{\text{specular component}}, \\
\bm{h} &= \text{normalize}(\bm{\omega}_o + \bm{\omega}_i),\\
F_0 &= (1 - M) \cdot 0.04 + M \cdot \bm{A}, \\
D(\bm{n}, \bm{h}) &= \frac{R^4}{\pi \left(
\bm{n} \cdot \bm{h} \ (R^4 - 1) + 1
\right)^2},\\
F(\bm{\omega}_i, \bm{n}) &= F_0 + (1 - F_0) \left(
1 - \bm{n} \cdot \bm{\omega}_i
\right)^5,\\
G(\bm{\omega}_o, \bm{\omega}_i, \bm{h}) &= G_1(\bm{\omega}_o, \bm{h}) \cdot G_1(\bm{\omega}_i, \bm{h}),\\
G_1(\bm{n}, \bm{h}) &= \frac{1}{1 + \bm{n} \cdot \bm{h} \sqrt{R^4 + \bm{n} \cdot \bm{h} - R^4 \cdot \bm{n} \cdot \bm{h}}},
\end{aligned}
\end{equation}
where $\bm{A}$, $R$, and $M$ denote the albedo, roughness, and metallicity, respectively. The Normal Distribution Function (NDF) $D$, Fresnel function $F$, and Geometry function $G$ are derived from physical materials. We use the Adam optimizer for training, and the training process is divided into three stages.

\begin{figure}[!th]
\includegraphics[width=\linewidth]{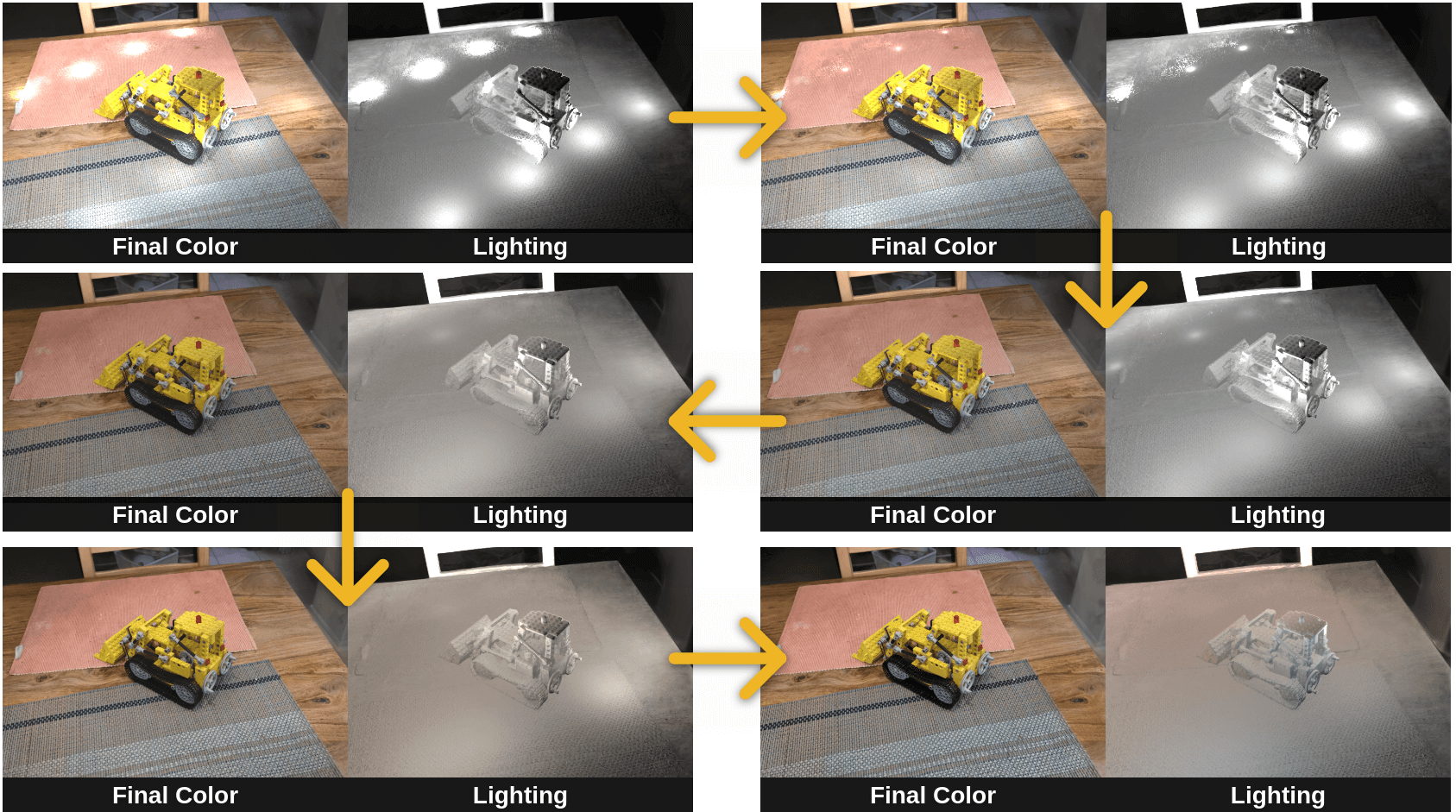}
\caption{We optimize for $N_{\text{light}}$ light sources with SGMs, in conjunction with a learnable environment map. Our representation is sufficiently expressive to capture detailed emissions while remaining controllable for light editing purposes.}
\label{fig:train_process}
\end{figure}

We employ a mixture model of a set of spherical Gaussians (SGMs) to represent the localized highlight illumination as $L^\text{SGMs}_{o}(\bm{x}, \bm{\omega}_o)$:
\begin{equation}
\begin{aligned}
L^\text{SGMs}_{o}(\bm{x}, \bm{\omega}_o) &= \sum_{i}^{N_\text{light}} \frac{{f_r}^\text{SGM}_{i} (\bm{n} \cdot \bm{\omega}_{i})*\textbf{V}_i}{d_i^2} \sum_{j}^{N_\text{sg}} W_{i, j} SG(j),
\end{aligned}
\end{equation}
where $d_i$ denotes the distance from the $i$-th spherical Gaussian mixture to the surface point $\bm{x}$. ${f_r}^\text{SGM}_{i}$ denotes the BRDF function, and $SG$ denotes the spherical Gaussian Function, respectively, as defined in the Methodology section. Each spherical Gaussian mixture contains $NSG$ spherical Gaussians. Utilizing pretrained 3DGS $\mathcal{T}$ and $\mathcal{C}$, along with supervision of material properties generated by a pretrained diffusion model, we optimize $M$ and $L$ over 15,000 iterations. The optimization is guided by the following loss function:
\begin{equation}
\begin{aligned}
\mathcal{L}_{\text{reg}} &= \lambda_{R} \text{L1}(R, \hat{R}) + \lambda_{M} \text{L1}(M, \hat{M}) + \lambda_{\bm{A}} \text{L1}(\bm{A}, \hat{\bm{A}}),
\end{aligned}
\label{eq:loss:material}
\end{equation}
where the weights for the loss function are set as: \(\lambda_{R} = 0.1\), \(\lambda_{M} = 0.1\), and \(\lambda_{\bm{A}} = 0.5\). The entire illumination optimization process can be visualized as shown in ~\Cref{fig:train_process}.

\section{Results on the MipNeRF 360, DB, and T\&T Datasets}
\begin{figure*}[t]
\includegraphics[width=1.0\textwidth]{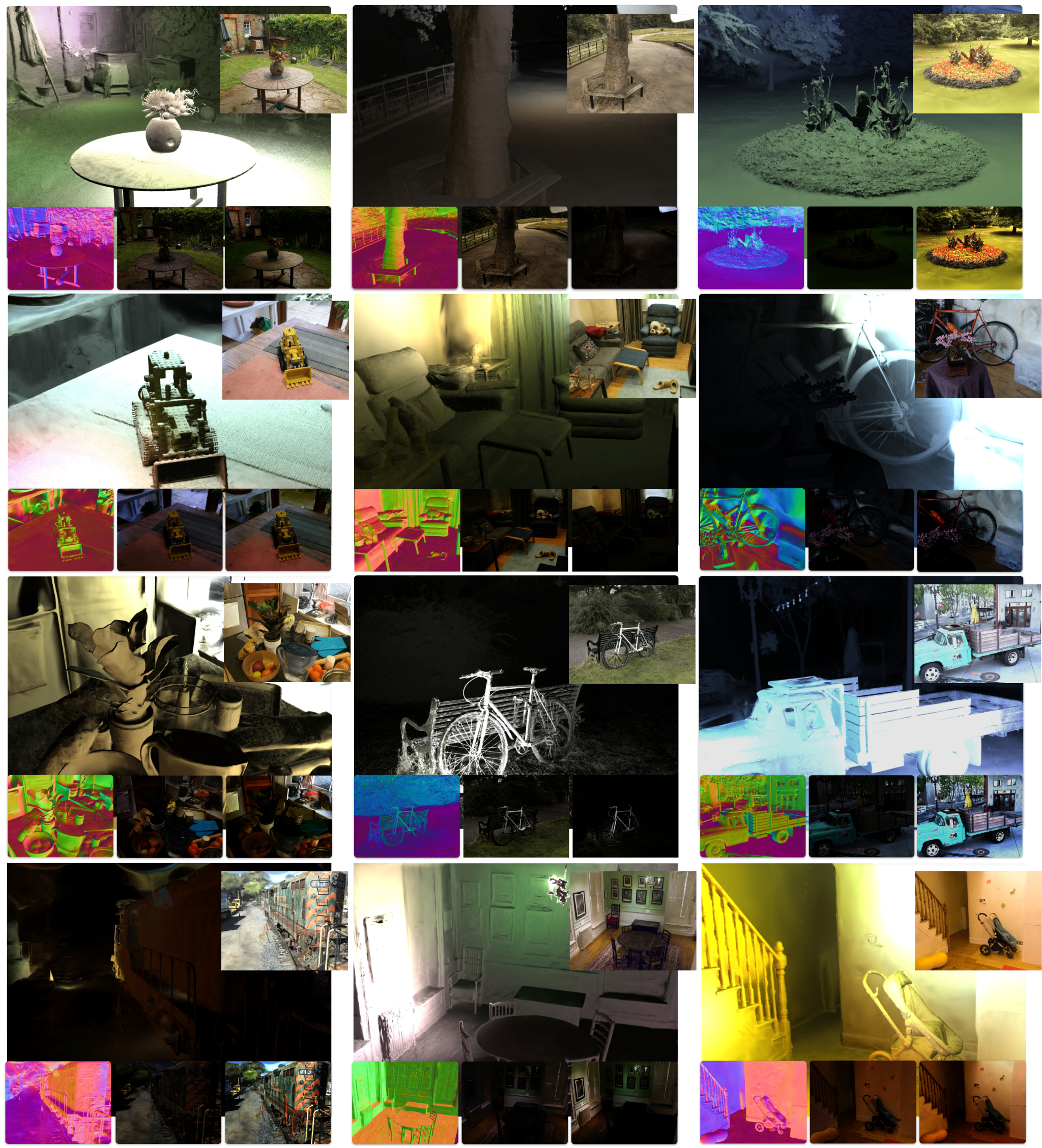}
\caption{Visualization of our illumination decomposition results on the Mip-NeRF 360, DeepBlending, and T\&T datasets. The main image shows the average of all effective pure localized light in the scene, while other smaller images include the normal map, environment light, localized light, and ambient occlusion.}
\label{fig:application_360}
\end{figure*}

Tabs.~\ref{tab:360psnr},~\ref{tab:360ssim}, and~\ref{tab:360lpips} present the results for novel view synthesis using the Mip-NeRF 360 Dataset. Tabs.~\ref{tab:ttdb_psnr},~\ref{tab:ttdb_ssim}, and~\ref{tab:ttdb_lpips} display the results for Deepblend and the T\&T Dataset. Additionally, ~\Cref{fig:application_360} illustrates the ID results for these scenes.

\section{Results on the TensoIR Synthetic Dataset}
~\Cref{tab:TensoIR} presents the outcomes for normal estimation, novel view synthesis, albedo reconstruction, and relighting across all four scenes.

\section{Results on the ADT Dataset}
~\Cref{tab:TensoIR} presents the outcomes for albedo estimation and novel view synthesis across all four scenes.

\begin{table}[!ht]
\renewcommand\arraystretch{1.3}
\centering
\scalebox{0.5}{
\begin{tabular}{l|c c c c c|c c c c}
Method & bicycle & flowers & garden & stump & treehill &
room & counter & kitchen & bonsai \\
\hline
NeRF++
& 22.64 & 20.31 & 24.32 & 24.34 & 22.20
& 28.87 & 26.38 & 27.80 & 29.15 \\
Plenoxels
& 21.91 & 20.10 & 23.49 & 20.66 & 22.25
& 27.59 & 23.62 & 23.42 & 24.67 \\
INGP-Base
& 22.19 & 20.35 & 24.60 & 23.63 & 22.36
& 29.27 & 26.44 & 28.55 & 30.34 \\
INGP-Big
& 22.17 & 20.65 & 25.07 & 23.47 & 22.37
& 29.69 & 26.69 & 29.48 & 30.69 \\
Mip-NeRF 360
& 24.40 & 21.64 & 26.94 & 26.36 & 22.81
& 29.69 & 26.69 & 29.48 & 30.69 \\
\hline
3DGS
& 25.25 & 21.52 & 27.41 & 26.55 & 22.49
& 30.63 & 28.70 & 30.32 & 31.98 \\
2DGS
&24.87& 21.15& 26.95& 26.47& 22.27& 31.06& 28.55& 30.50& 31.52 \\
\hline
GaussianShader
&23.12&20.34&26.44&23.92&20.17&24.27&26.35&27.72&28.09 \\
GSIR
& 23.80 &20.57& 25.72& 25.37 &21.79
& 28.79 &26.22 &27.99& 28.18 \\
Ours ($3^3$ SGMs)
& 24.41 & 20.90 & 26.76 & 24.35 & 22.01
& 30.47 & 27.42 & 29.95 & 30.18 \\
\end{tabular}
}
\caption{PSNR scores for Mip-NeRF360 scenes.}
\vspace{-0.4cm}
\label{tab:360psnr}
\end{table}

\begin{table}[!ht]
\renewcommand\arraystretch{1.3}
\centering
\scalebox{0.5}{
\begin{tabular}{l|c c c c c|c c c c}
Method & bicycle & flowers & garden & stump & treehill &
room & counter & kitchen & bonsai \\
\hline
NeRF++
& 0.526 & 0.453 & 0.635 & 0.594 & 0.530
& 0.530 & 0.802 & 0.816 & 0.876 \\
Plenoxels
& 0.496 & 0.431 & 0.606 & 0.523 & 0.509
& 0.842 & 0.759 & 0.648 & 0.814 \\
INGP-Base
& 0.491 & 0.450 & 0.649 & 0.574 & 0.518
& 0.855 & 0.798 & 0.818 & 0.890 \\
INGP-Big
& 0.512 & 0.486 & 0.701 & 0.594 & 0.542
& 0.871 & 0.817 & 0.858 & 0.906 \\
Mip-NeRF 360
& 0.693 & 0.583 & 0.816 & 0.746 & 0.632
& 0.913 & 0.895 & 0.920 & 0.939 \\
\hline
3DGS
&0.771& 0.605& 0.868& 0.775& 0.638& 0.914& 0.905& 0.922& 0.938 \\
2DGS
& 0.752 &0.588& 0.852& 0.765& 0.627& 0.912& 0.900& 0.919& 0.933  \\
\hline
GaussianShader
& 0.700 &	0.542& 0.842& 0.667& 0.572& 0.847& 0.874& 0.887& 0.893 \\
GSIR
& 0.706& 0.543& 0.804 &0.716& 0.586 &0.867 &0.839 &0.867 &0.883 \\
Ours ($3^3$ SGMs)
& 0.734 & 0.561 & 0.826 & 0.665 & 0.586
& 0.906 & 0.883 & 0.910 & 0.923 \\
\end{tabular}
}
\caption{SSIM scores for Mip-NeRF360 scenes.}
\label{tab:360ssim}
\end{table}

\begin{table}[!ht]
\renewcommand\arraystretch{1.3}
\centering
\scalebox{0.5}{
\begin{tabular}{l|c c c c c|c c c c}
Method & bicycle & flowers & garden & stump & treehill &
room & counter & kitchen & bonsai \\
\hline
NeRF++
& 0.455 & 0.466 & 0.331 & 0.416 & 0.466
& 0.335 & 0.351 & 0.260 & 0.291 \\
Plenoxels
& 0.506 & 0.521 & 0.386 & 0.503 & 0.540 
& 0.419 & 0.441 & 0.447 & 0.398 \\
INGP-Base
& 0.487 & 0.481 & 0.312 & 0.450 & 0.489
& 0.301 & 0.342 & 0.254 & 0.227 \\
INGP-Big
& 0.446 & 0.441 & 0.257 & 0.421 & 0.450
& 0.261 & 0.306 & 0.195 & 0.205 \\
Mip-NeRF 360
& 0.289 & 0.345 & 0.164 & 0.254 & 0.338
& 0.211 & 0.203 & 0.126 & 0.177 \\
\hline
3DGS
&0.205& 0.336& 0.103& 0.210& 0.317& 0.220& 0.204& 0.129& 0.205 \\
2DGS
& 0.218& 0.346& 0.115& 0.222& 0.329& 0.223& 0.208& 0.133& 0.214 \\
\hline
GaussianShader
& 0.274 	&0.377& 	0.130& 	0.297& 	0.406& 	0.304& 	0.242& 	0.167& 	0.257  \\
GSIR
& 0.259 &0.371 &0.158 &0.258 &0.372 &0.279 &0.260 &0.188 &0.264 \\
Ours ($3^3$ SGMs)
& 0.249 & 0.368 & 0.144 & 0.329 & 0.400
& 0.235 & 0.229 & 0.146 & 0.226 \\
\end{tabular}
}
\caption{LPIPS scores for Mip-NeRF360 scenes.}
\vspace{-0.3cm}
\label{tab:360lpips}
\end{table}

\begin{table}[!ht]
\renewcommand\arraystretch{1.3}
\centering
\scalebox{0.7}{
\begin{tabular}{l|c c c | c c c }
Method & Trucks & Train & Avg. & Drjohnson & Playroom & Avg. \\
\hline
3DGS
& 25.18& 21.09 & 23.13 & 28.76 &30.04&29.40 \\
2DGS
& 24.78 & 21.67 & 23.22 & 29.02 & 30.23&29.62 \\
\hline
GaussianShader
& 20.13 & 23.56 &21.84& 22.18 & 19.88 & 21.03 \\
GSIR
& 24.09 & 20.42 & 22.25 & 26.47 & 28.13 &27.30\\
Ours ($3^3$ SGMs)
& 24.51 & 23.60 & 22.71 & 28.05 & 28.55 & 28.30\\
\end{tabular}
}
\caption{PSNR scores for DB and T\&T Dataset.}
\vspace{-0.3cm}
\label{tab:ttdb_psnr}
\end{table}

\begin{table}[!ht]
\renewcommand\arraystretch{1.3}
\centering
\scalebox{0.7}{
\begin{tabular}{l|c c c | c c c }
Method & Trucks & Train & Avg. & Drjohnson & Playroom & Avg. \\
\hline
3DGS
& 0.879 & 0.802 & 0.840 & 0.899 & 0.906 & 0.902  \\
2DGS
& 0.867 & 0.803 & 0.835 & 0.901 & 0.907 & 0.904 \\
\hline
GaussianShader
& 0.763 & 0.843 & 0.803 & 0.786 & 0.845 &0.815 \\
GSIR
& 0.833 & 0.742 & 0.787 & 0.863 & 0.869 & 0.866 \\
Ours ($3^3$ SGMs)
& 0.858 & 0.778 & 0.818 & 0.886 & 0.885 & 0.885 \\
\end{tabular}
}
\caption{SSIM scores for DB and T\&T Dataset.}
\label{tab:ttdb_ssim}
\end{table}

\begin{table}[!ht]
\renewcommand\arraystretch{1.3}
\centering
\scalebox{0.7}{
\begin{tabular}{l|c c c | c c c }
Method & Trucks & Train & Avg. & Drjohnson & Playroom & Avg. \\
\hline
3DGS
& 0.148 &0.218 & 0.183 &0.244 &0.241 & 0.242  \\
2DGS
& 0.183 & 0.227 & 0.205 &  0.248 & 0.250 & 0.249  \\
\hline
GaussianShader
& 0.271 & 0.191 & 0.231 & 0.336 & 0.359 & 0.344\\
GSIR
& 0.195 & 0.273 & 0.234 & 0.314 & 0.305 & 0.309  \\
Ours ($3^3$ SGMs)
& 0.182 & 0.250 &0.216  & 0.268 & 0.271&0.269 \\
\end{tabular}
}
\caption{LPIPS scores for DB and T\&T Dataset.}
\label{tab:ttdb_lpips}
\end{table}

\begin{table*}[!ht]
\centering
\scalebox{0.99}{
\begin{tabular}{l c|c|c c c|c c c}
&&{{Normal (MAE)} } &
\multicolumn{3}{c|}{Albedo} &
\multicolumn{3}{c}{Novel View Synthesis} \\
& & &
PSNR & SSIM  & LPIPS  &
PSNR & SSIM  & LPIPS  \\
\hline
{Lego}
& InvRender & 9.980 &
21.435 & 0.882 & 0.160 &
24.391 & 0.883 & 0.151 \\
& TensoIR   & 5.980  &
25.240 & 0.900 & 0.145 &
34.700 & 0.968 & 0.037 \\
& GS-IR      & 8.078  &
24.958 & 0.889 & 0.143 &
34.379 & 0.968 & 0.036 \\
& Ours ($3^3$ SGMs)      & 7.595  &
25.155 & 0.911 & 0.138 &
37.377 & 0.975 & 0.023 \\
\hline
{Hotdog}
& InvRender & 3.708 &
27.028 & 0.950 & 0.094 &
31.832 & 0.952 & 0.089 \\
& TensoIR   & 4.050 &
30.370 & 0.947 & 0.093 &
36.820 & 0.976 & 0.045 \\
& GS-IR      & 4.771 &
26.745 & 0.941 & 0.088 &
34.116 & 0.972 & 0.049 \\
& Ours ($3^3$ SGMs)    & 4.269 &
31.546&0.949&0.0874&
37.599 & 0.983 & 0.024 \\
\hline
{Armadillo}
& InvRender & 1.723 &
35.573 & 0.959 & 0.076 &
31.116 & 0.968 & 0.057 \\
& TensoIR   & 1.950 &
34.360 & 0.989 & 0.059 &
39.050 & 0.986 & 0.039 \\
& GS-IR      & 2.176 &
38.572 & 0.986 & 0.051 &
39.287 & 0.980 & 0.039 \\
& Ours ($3^3$ SGMs)     & 2.927 &
44.585&0.986&0.044 &
43.858 & 0.988 & 0.024 \\
\hline
{Ficus}
& InvRender & 4.884 &
25.335 & 0.942 & 0.072 &
22.131 & 0.934 & 0.057 \\
& TensoIR   & 4.420 &
27.130 & 0.964 & 0.044 &
29.780 & 0.973 & 0.041 \\
& GS-IR      & 4.762 &
30.867 & 0.948 & 0.053 &
33.551 & 0.976 & 0.031 \\
& Ours ($3^3$ SGMs)    & 3.681 &
32.702&0.959&0.049 &
37.681 & 0.990 & 0.008 \\
\hline
\end{tabular}
}
\caption{Per-scene results on TensoIR Synthetic dataset. For albedo reconstruction results, we follow NeRFactor and scale each RGB channel using a global scalar.}
\label{tab:TensoIR}
\vspace{-0.5cm}
\end{table*}

\begin{table*}[!ht]
\centering
\scalebox{0.99}{
\begin{tabular}{l c|c c c|c c c}
&&
\multicolumn{3}{c|}{Albedo} &
\multicolumn{3}{c}{Novel View Synthesis} \\
& & 
PSNR & SSIM  & LPIPS  &
PSNR & SSIM  & LPIPS  \\
\hline
{Donut}
& GS-IR & 
25.224& 	0.949& 	0.029 &
35.171& 	0.975& 	0.046\\
& GSshader & 
24.138&	0.941&	0.031&
34.171&	0.971&	0.047 \\
& RelightGS   & 
27.315&	0.959&	0.025& 
37.414&	0.988&	0.019 \\
& Ours ($3^3$ SGMs)  & 
28.515&	0.969&	0.0216 &
38.414&	0.989&	0.018 \\
\hline
{Figurine}
& GS-IR & 
25.224& 	0.949& 	0.029 &
35.171& 	0.975& 	0.046\\
& GSshader & 
28.563&	0.905&	0.059 &
37.709&	0.961&	0.048 \\
& RelightGS   & 
31.295& 	0.972& 	0.046& 
41.933&	0.992&	0.011 \\
& Ours ($3^3$ SGMs)  & 
30.895 &0.963 &	0.048 &
40.633&	0.986&	0.012 \\
\hline
{Birdhouse}
& GS-IR & 
26.584 &0.906 &	0.121 &
35.814&	0.972&	0.035\\
& GSshader & 
23.584&	0.890&	0.125&
34.991&	0.968&	0.039 \\
& RelightGS   & 
28.011&  	0.933& 	0.094& 
38.287&	0.986&	0.012 \\
& Ours ($3^3$ SGMs)  & 
26.084& 	0.926&	0.101&
38.021&	0.982&	0.013 \\
\hline
{Table}
& GS-IR & 
25.718&	0.938&	0.063 &
38.439&	0.972&	0.036\\
& GSshader & 
25.213&0.912&	0.064 &
38.123&	0.969&	0.038\\
& RelightGS   & 
27.132&	0.948&	0.062& 
43.280& 0.996& 	0.006 \\
& Ours ($3^3$ SGMs)  & 
29.752&	0.958&	0.055 &
44.581&	0.994&	0.006 \\
\hline
\end{tabular}
}
\caption{Per-scene results on ADT dataset. For albedo reconstruction results, we follow NeRFactor and scale each RGB channel using a global scalar.}
\label{tab:ADT}
\vspace{-0.5cm}
\end{table*}

\section{Applications}

\begin{figure*}[t]
\centering
\includegraphics[width=0.92\textwidth]{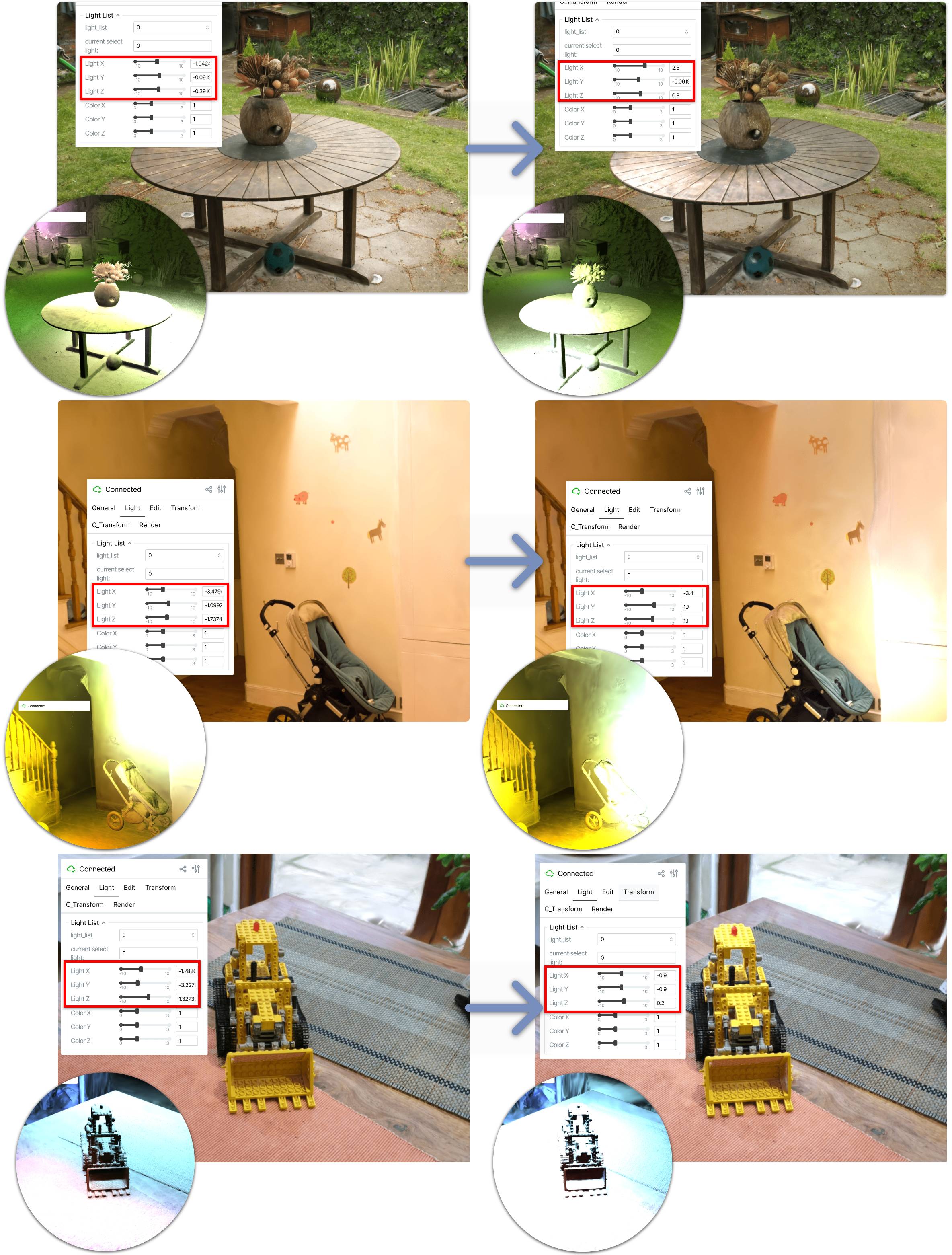}
\caption{Visualization of light editing results.}
\label{fig:application_le}
\end{figure*}

\textbf{Light Editing}
The use of a spherical Gaussian (SG) mixture for lighting provides a flexible and intuitive representation of scene illumination. After training, the emission weights, positions, and SG parameters of light sources can be independently modified. This explicit emissive formulation enables physically plausible and interactive light editing, as shown in \Cref{fig:application_le} and the accompanying video.

\begin{figure*}[t]
\centering
\includegraphics[width=0.73\textwidth]{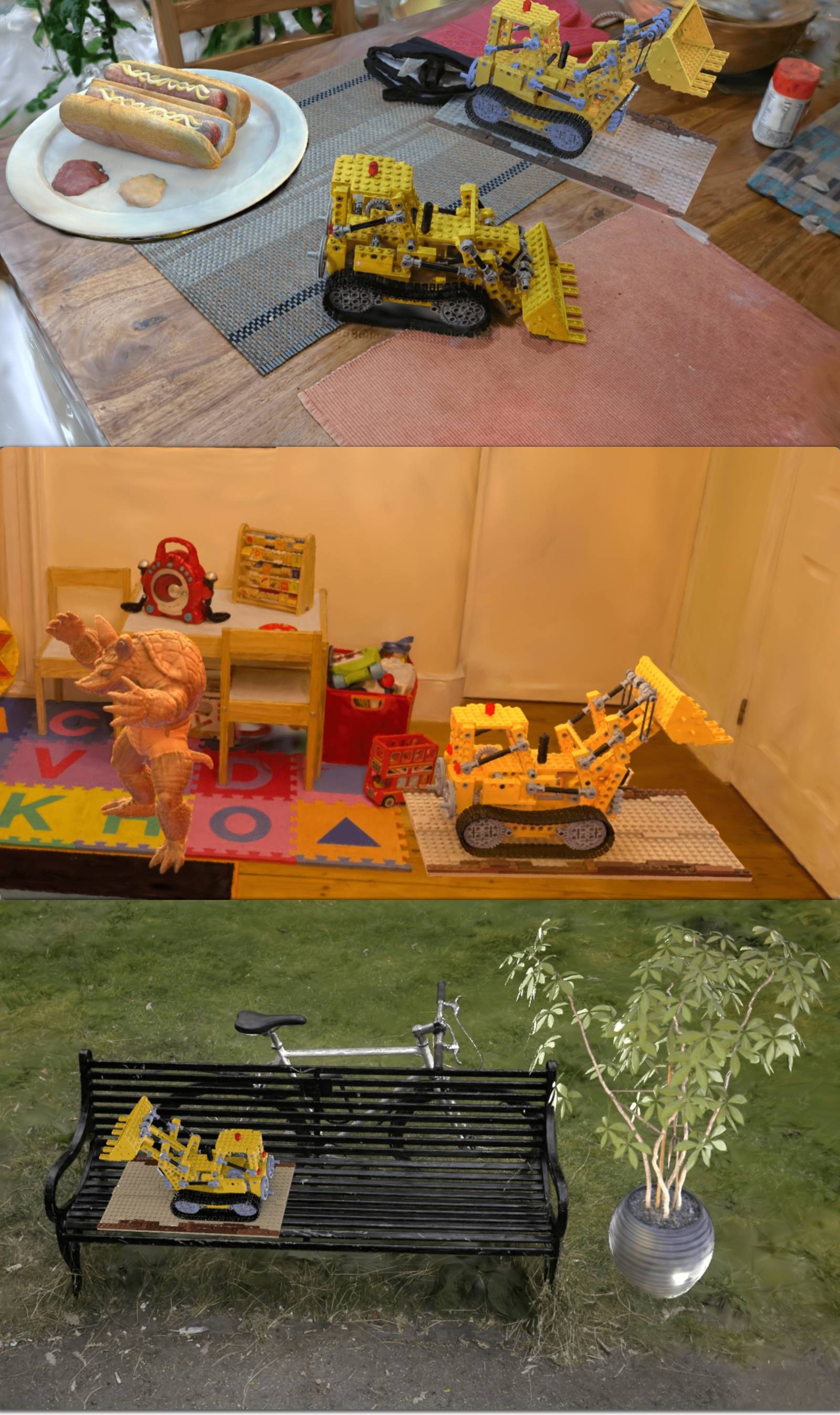}
\caption{Visualization of scene composition results.}
\label{fig:application_com}
\end{figure*}

\textbf{Composition}
Our method models the scene's illumination as a combination of localized and environment lighting, allowing for complete illumination decomposition. This decomposition enables seamless integration of relightable content across different scenes. As demonstrated in \Cref{fig:application_com} and the video, we incorporate a decomposed TensoIR-synthesized object into a real-world Mip-NeRF scene, achieving realistic lighting consistency.

\begin{figure*}[t]
\includegraphics[width=1.0\textwidth]{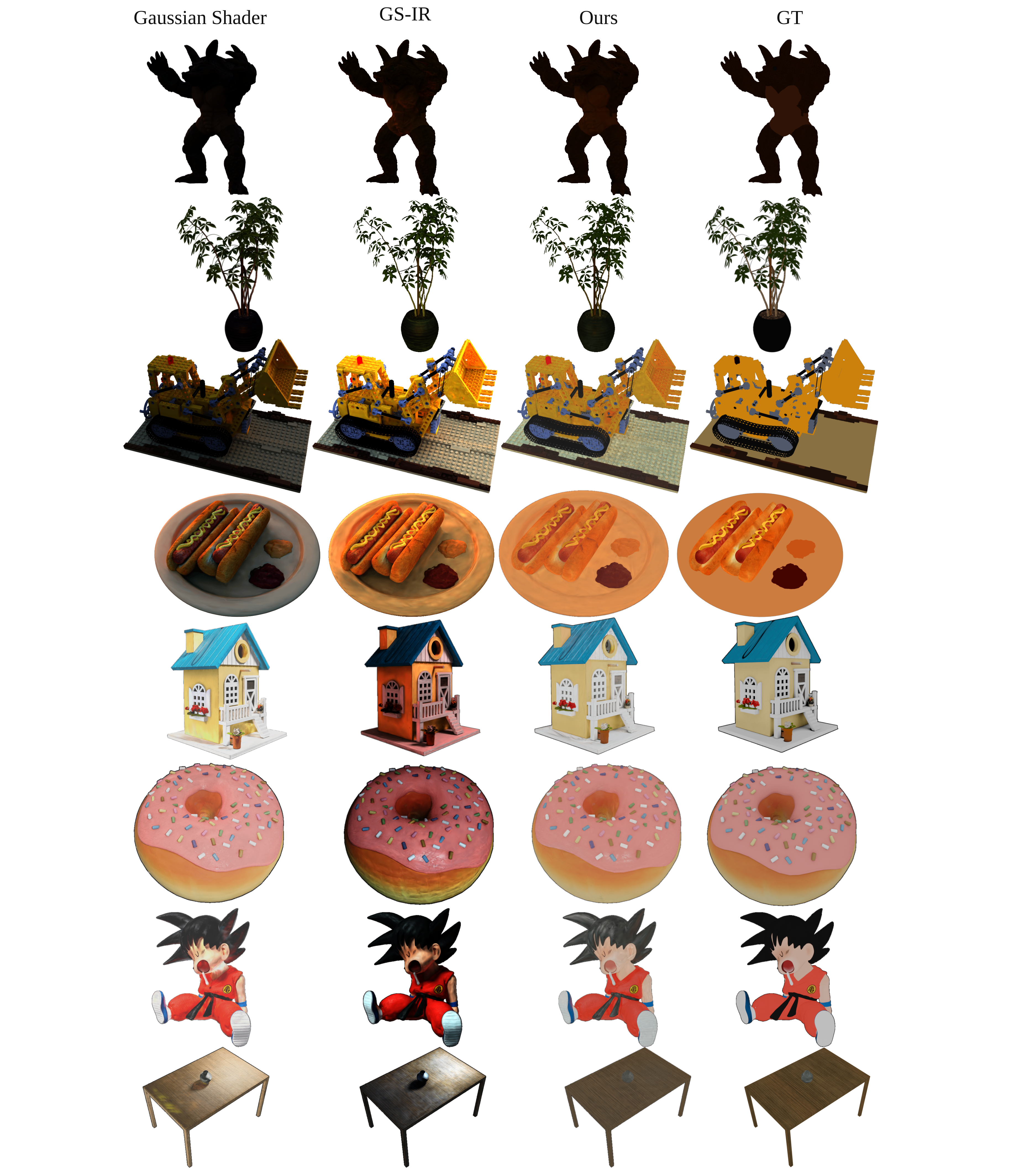}
\caption{Visualization of our albedo estimation results with other methods on the TensoIR synthetic and ADT dataset.}
\label{fig:application_albedo}
\end{figure*}

\end{document}